\documentclass[10pt,twocolumn,letterpaper]{article}

\usepackage[pagenumbers]{cvpr} 

\definecolor{cvprblue}{rgb}{0.21,0.49,0.74}
\usepackage[pagebackref,breaklinks,colorlinks,allcolors=cvprblue]{hyperref}

\usepackage{subcaption}
\usepackage{booktabs} 
\usepackage{graphicx}
\usepackage{amsmath}
\usepackage{amssymb}
\usepackage{mathtools}
\usepackage{amsthm}
\usepackage{wrapfig}
\usepackage{adjustbox}
\usepackage{booktabs}
\usepackage{multirow}
\usepackage[table,xcdraw]{xcolor}

\def\paperID{276} 
\def\confName{CVPR}
\def\confYear{2026}

\title{Resilient Contrastive Pre-training under Non-Stationary Drift}

\author{Xiaoyu Yang, Jie Lu, En Yu, Wei Duan \\
Australian Artificial Intelligence Institute (AAII), \\ 
Faulty of Engineering and Information Technology,\\
University of Technology Sydney, Australia.\\
}

\begin{document}
\maketitle

\begin{abstract}


The remarkable success of large-scale contrastive pre-training has been largely driven by by vast yet static datasets. However, as the scaling paradigm evolves, this paradigm encounters a fundamental challenge when applied to dynamic data streams characterized by concept drift—unpredictable changes in the underlying data distribution. This paper aims to advance robust pre-training under such non-stationary environments. We begin by revealing that conventional contrastive pre-training methods are highly susceptible to concept drift, resulting in significant substantial bias and instability within the learned feature representations. To systematically analyze these effects, we develop a structural causal model that elucidates how drift acts as a confounder, distorting the learned representations. Based on these causal insights, we propose \textbf{R}esilient \textbf{C}ontrastive \textbf{P}re-training (RCP), a novel method that incorporates causal intervention. RCP formulates a causally-informed objective to mitigate drift-induced biases through targeted interventions. The method is designed for simple and scalable implementation and exhibits notable adaptability, promoting robust and autonomous pre-training on non-stationary data. Comprehensive experiments across various downstream tasks consistently demonstrate that RCP effectively alleviates the detrimental impact of concept drift, yielding more resilient and generalizable representations. Codes are available at \url{https://anonymous.4open.science/r/ResilientCL/}.

\end{abstract}

\section{Introduction}


Contrastive learning has emerged as a dominant paradigm for large-scale visual pre-training, demonstrating impressive performance across a wide range of vision tasks~\cite{heMomentumContrastUnsupervised2020, caronEmergingPropertiesSelfSupervised2021, zhouDINOWMWorldModels2024}. While most existing studies evaluate contrastive methods on static, fully-scraped internet corpora, the pre-training landscape is shifting toward dynamic and continuously expanding data regimes\cite{yang2024adaptingmultimodallargelanguage,yang2025learning}. Modern large-scale systems no longer rely on a one-shot frozen dataset. Instead, they accumulate data over time from heterogeneous sources such as user-generated content in federated/edge-learning pipelines\cite{gao2022feddc,chenimportance,panchal2023flash}, medical imaging networks \cite{jiang2022harmofl,chen2024general,yang2022local,yang2024segmentation,yang2021variational}, and autonomous-driving fleets \cite{liang2022effective,zheng2023autofed}. In these real-world settings, data arrive sequentially, exhibit strong non-IID characteristics, and naturally undergo concept drift, including evolving class frequencies, shifting domains, seasonal changes, and sensor-induced feature heterogeneity~\cite{luLearningConceptDrift2019,yang2024adaptingmultimodallargelanguage,yang2025walking}. Recent studies in federated learning~\cite{chenimportance,jiang2022harmofl,panchal2023flash} and autonomous driving~\cite{zheng2023autofed,liang2022effective} consistently highlight that drifting data distributions degrade representation quality and destabilize pre-trained backbones. These developments raise a critical but under-explored question: 

\textbf{\textit{Can contrastive pre-training reliably learn from non-stationary and drifting data streams?}}

Through theoretical analysis and empirical evidence, we show that current contrastive objectives are highly sensitive to temporal distribution shift, leading to biased momentum teachers, collapsed similarity structures, and unstable representation dynamics. This exposes an urgent need for principled approaches that make contrastive pre-training robust to drift in realistic, continuously evolving environments. A review of related work is provided in the Appendix.


Current contrastive pre-training methods primarily rely on maximizing the agreement between two augmented views of the same input instance, typically processed through different encoders or network states~\cite{chenSimpleFrameworkContrastive2020, grillBootstrapYourOwn2020, heMomentumContrastUnsupervised2020}. In the context of large vision models, this paradigm is often instantiated through student–teacher architectures, as exemplified by frameworks such as DINO~~\cite{caronEmergingPropertiesSelfSupervised2021} and MoCo v3~\cite{chenEmpiricalStudyTraining2021}. In these setups, the student network is optimized via a contrastive objective, while the teacher network is updated using an exponential moving average (EMA) of the student parameters to stabilize training.
However, this widely adopted momentum-based update mechanism becomes particularly vulnerable under concept drift. As the underlying data distribution evolves, the teacher network—updated only gradually through EMA—tends to retain outdated knowledge from previous distributions.
This stale information accumulates over time, propagating outdated representations and amplifying the misalignment between the learned feature space and the current data characteristics, ultimately degrading the quality of the pre-trained representations.

\begin{figure}[htbp]
    \centering
    \begin{subfigure}[t]{0.23\textwidth}
        \centering
        \includegraphics[height=0.15\textheight]{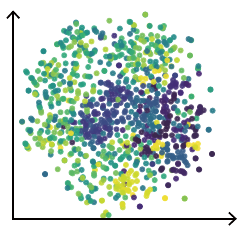}
        \caption{Balanced Pre-training}
        \label{fig:balanced}
    \end{subfigure}
    \begin{subfigure}[t]{0.23\textwidth}
        \centering
        \includegraphics[height=0.15\textheight]{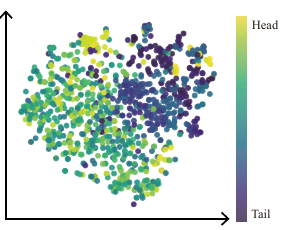}
        \caption{Tailed Drift Pre-training}
        \label{fig:imbalanced}
    \end{subfigure}
    \caption{The t-SNE visualization of feature space under the different conditions of pre-training within ImageNet and ImageNet-LT. The dark colors signify the region corresponding to the tail category with limited pre-training samples, whereas light colors denote the head category characterized by abundant samples.}
    \label{fig:analysis}
\end{figure}

To concretely illustrate this phenomenon, we examine the effect of long-tailed drift—a prevalent form of concept drift where class frequencies evolve over time—on the contrastive pre-training process. We visualize the feature space learned by MoCo v3~\cite{chenEmpiricalStudyTraining2021} in Figure~\ref{fig:analysis}. We simulate the pre-training process on two scenarios: (a) standard pre-training on the balanced ImageNet dataset~\cite{russakovskyImagenetLargeScale2015}; and (b) drift pre-training on ImageNet-LT~\cite{liuLargeScaleLongTailedRecognition2019}, which simulates a non-stationary stream with an increasingly imbalanced distribution. 
For a fair comparison, the feature embeddings are projected using t-SNE and evaluated on a balanced test set. Category colors range from dark to light, representing the transition from tail classes (few training samples) to head classes (abundant samples). As shown in Figure~\ref{fig:analysis}, the feature space under long-tailed drift becomes highly distorted: tail categories collapse into a compact region, while head categories dominate most of the embedding manifold.
This imbalance contrasts sharply with the well-structured and evenly distributed feature space obtained under balanced pre-training.
Such degradation can be attributed to the bias accumulation in the momentum-updated teacher network, which is repeatedly exposed to head-class samples.
As the distribution drifts, the teacher provides outdated or skewed supervision, leading to suboptimal representation learning for the under-represented classes.

Building the above analysis, we formalize the interplay among these components using a Structural Causal Model (SCM), illustrated in Figure~\ref{fig:causal_model}. It elucidates the causal relationships among input samples $X$, predictions $Y$, the latent concept drift 
$D$ inherent in the data stream, and the bias $B$ accumulated in the momentum-updated teacher network~\cite{pearl2014interpretation}.
Our causal analysis (detailed in Section~\ref{section:causal}) reveals that concept drift $D$ acts as a critical confounder. It not only influences the input features $X$ ($D \rightarrow X$) but also shapes the bias $B$ within the teacher network $(X, D) \rightarrow B$, which in turn affects $Y$, i.e., $(X, B) \rightarrow Y$. This induces a spurious backdoor path $X\leftarrow D \rightarrow B \rightarrow Y$ that can mislead the learning process by inducing false correlations between $X$ and $Y$. Furthermore, the mediation~\cite{pearl2022direct} path $X \rightarrow B \rightarrow Y$ confounds the direct contribution of $X \rightarrow Y$, as detailed in Section~\ref{section:causal}. 

These causal insights underscore the limitations of standard contrastive methods in drifting environments. Therefore, to counteract the detrimental effects of confounding and biased mediation, we propose a novel pre-training method, \textbf{R}esilient \textbf{C}ontrastive \textbf{P}re-training (\textbf{RCP}).
RCP leverages causal intervention techniques~\cite{pearl2016causal} to actively mitigate the bias $B$ propagated by concept drift $D$, thereby enabling more robust and generalizable representation learning from non-stationary data streams.
In summary, our paper mainly makes the following contributions:
\begin{enumerate} 
    
    \item We are pioneers in uncovering how concept drift impacts contrastive pre-training, specifically identifying the accumulation of bias through the momentum update mechanism. This foundational understanding paves the way for future research into robust pre-training on large-scale, dynamic datasets.
    
    \item Leveraging insights from our structural causal model, we propose resilient contrastive pre-training (RCP), a novel method that employs causal intervention to mitigate drift-induced bias. RCP is designed to be straightforward to implement and scalable, making it suitable for pre-training large models on evolving data streams.

    \item Extensive experiments demonstrate the superiority of RCP across diverse downstream tasks, e.g., long-tailed classification, OOD detection, and domain adaptation, using both fine-tuning and linear probing evaluation protocols. Notably, our analysis of inter-category distances in the learned feature space confirms that RCP effectively mitigates the detrimental effects of drift, leading to more discriminative and robust representations for large-scale pre-training.

\end{enumerate}

\section{Methodology}

\subsection{Causal Lens on Contrastive Drift Pre-training}
\label{section:causal}
\textbf{Contrastive Drift Pre-training (CDP).} 
Concept drift is a statistical phenomenon wherein the joint distribution $P(X,Y)$ of features $X$ and target variables $Y$ evolves over time~\cite{luLearningConceptDrift2019}. We define Drift Pre-training as a pre-training paradigm operating on a data stream $\mathcal{S}_{0,T} = \{S_0, S_1, \dots, S_T\}$.
Each element $S_t=\{(x_i,y_i)\}_{i=1}^{n_t}$ consists of ${n_t}$ samples with feature vector $x_i$ and an associated contrastive label $y_i$ at timestamp $t$. This timestamp $t$ corresponds to an iteration in the pre-training process.
At any given iteration $t$, data samples $\{X, Y\} \in \{(x_i,y_i)\}_{i=1}^{n_t}$ are drawn from a joint distribution $P_t(X, Y)$. Concept drift within this pre-training context is then formally characterized by:
\begin{equation} 
\label{eq:concept_drift_formalism}
    \exists t \in [0, T) \text{:} P_t(X, Y) \neq P_{t+1}(X, Y),
\end{equation}
where the joint probability $P_{t}(X, Y)$ can be decomposed as $P_{t}(X, Y) = P_{t}(X)\times P_{t}(Y | X)$. Specifically, tailed drift (or long-tailed imbalance) primarily reflects variations in the marginal distribution $P_t(X)$, where the frequency of certain classes or patterns changes over time.
In contrast, out-of-distribution (OOD) drift corresponds to alterations in the conditional distribution $P_t(Y|X)$, indicating that the mapping from features to labels evolves with new or redefined concepts. Both phenomena are manifestations of concept drift, where either the data distribution or the underlying labeling function (or both) changes unpredictably over time. In this work, we aim to develop a pre-training paradigm that can explicitly account for and adapt to these coupled forms of drift, enabling contrastive models to remain robust and consistent across dynamic environments.


CDP investigates contrastive learning under such non-stationary conditions, typically employing momentum-updated teacher-student frameworks~\cite{heMomentumContrastUnsupervised2020}, i.e., 
\begin{equation}
\label{eq:ema_cdp} 
    \theta^{m}_{t} = \lambda\theta^{m}_{t-1} +(1-\lambda) \theta^{g}_{t-1},
\end{equation}
where $\theta^{m}$ signifies the parameters of the teacher network, $\theta^{g}$ denotes those of the student network, and $\lambda\in [0,1)$ represents the momentum coefficient. Momentum $ \lambda\theta^{m}_{t-1}$ induces a slower evolution of the teacher network $\theta^{m}$ relative to the student network $\theta^{s}$. While high momentum (e.g., $\lambda=0.999$~\cite{chenEmpiricalStudyTraining2021}) stabilizes training on stationary data, it poses a significant challenge under concept drift: the slowly updating teacher network $\theta^m$ can become severely misaligned with the current data distribution $P_t(x,z)$, thereby accumulating stale biases from past distributions and hindering the model's ability to adapt to new or evolved concepts present in the current iteration $S_t$.
\begin{figure}
    \centering
    \begin{subfigure}[t]{0.23\textwidth}
        \centering
      \includegraphics[width=0.99\textwidth]{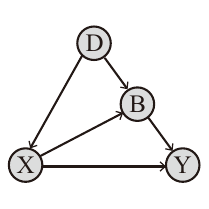}
        \caption{SCM}
        \label{fig:causal_model}
    \end{subfigure}
    \begin{subfigure}[t]{0.23\textwidth}
        \centering
        \includegraphics[width=0.99\textwidth]{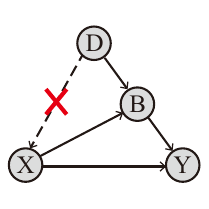}
        \caption{Intervention}
        \label{fig:intervention}
    \end{subfigure}
    \caption{The proposed causal graph of contrastive pre-training. \textbf{X}: Sample Features, \textbf{Y}: Prediction, \textbf{D}: Latent Concept Drift within Data Streams, and \textbf{B}: Sample Bias in the Momentum Update.}
    \label{fig:causal}
\end{figure}

\textbf{Objective.} 
Therefore, CDP aims to enable the learning system to synchronize with the evolving data distribution, ensuring that the pre-training model accurately reflect the current state of the data. We formalize this objective as maximizing the agreement between the student model and momentum model over a prospective adaptation window $[t, t+\tau]$:
\begin{equation}
    \label{eq:max}
     \max_{\{g_{j}^{\theta}\}_{j=t}^{t+\tau}} \sum_{j=t}^{t+\tau} \mathcal{L}_{\text{agree}}(g_{j}^{\theta}(\tilde{X}_{j}), m_{j}^{\theta}(\hat{X}_{j}))
\end{equation}
where $\tilde{X}_{j}$ and $\hat{X}_{j}$ symbolize different augmented samples obtained from ${X}_{j}$, $g_{j}^{\theta}$ denotes the student model trained by the data stream $S_{j-\tau, j-1}$ from the drift adaption window with the size of $\tau$, and $m_{j}^{\theta}$ represents the momentum model. And the pre-training is driven by the target metric $\mathcal{L}_{\text{agree}}$ to continuously adapt the drift in a given drift adaptation window of $[t, t+\tau]$ time period.
This is a standard objective for continuous contrastive learning but fails to explicitly handle the accumulated bias $B$ caused by drift $D$, which we formalize next.

\textbf{Causal Analysis of CDP.}
To further understand the challenges posed by concept drift in CDP, we employ a Structural Causal Model (SCM)~\cite{pearl1995causal, pearl2016causal}.
As shown in Figure~\ref{fig:causal}, the constructed causal graph of $\{X,Y,D,B\}$ presents the following causal connections:

$D\rightarrow X$: $X$ denotes features extracted by samples drawn from the data stream with concept drift $D$, which is obviously trained under the effect of the drift pre-training.  

$(X, D) \rightarrow B$: $ B$ represents the sample bias deviated from feature $ X$ under the effect of the concept drift $ D$ within the data stream. In the context of drift pre-training, the bias will accumulate during the momentum update of the teacher network, which will be amplified in the subsequent iteration of the contrastive pre-training.

$ (X, B) \rightarrow Y$: This link presents that, apart from the regular $X \rightarrow Y$, the prediction is also impacted by the concept drift within the data stream through the mediation bias of $B$.


This structure reveals two detrimental pathways that compromise the true causal link $X \rightarrow Y$:

\noindent
\textbf{Backdoor Path (Spurious Correlation):} $X\leftarrow D\rightarrow B \rightarrow Y$. Concept drift $D$ acts as a confounder. This path introduces a spurious correlation, meaning $X$ and $Y$ appear related solely due to their mutual dependency on $D$ and $B$. For instance, in long-tailed drift, tail samples might be spuriously associated with head categories due to the overwhelming head-class bias $B$.

\noindent
\textbf{Mediation Path (Weakening Effect):} $X\rightarrow B \rightarrow Y$. The bias $B$ acts as a mediator. This path conveys the drift effect, effectively capturing and weakening the direct and desirable impact of the true feature $X$ on the prediction $Y$.

\begin{figure*}[htbp]
    \centering
    \begin{subfigure}[t]{0.19\textwidth}
        \centering
        \includegraphics[height=.22\textheight]{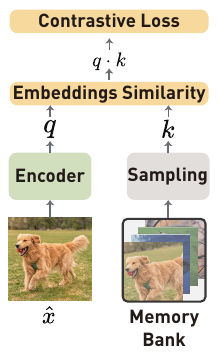}
        \caption{Memory Bank}
        \label{fig:cl}
    \end{subfigure}
    \begin{subfigure}[t]{0.19\textwidth}
        \centering
        \includegraphics[height=.22\textheight]{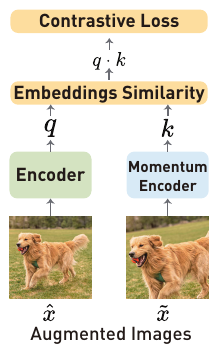}
        \caption{MoCo}
        \label{fig:mim}
    \end{subfigure}
    \begin{subfigure}[t]{0.6\textwidth}
        \centering
        \includegraphics[height=.23\textheight]{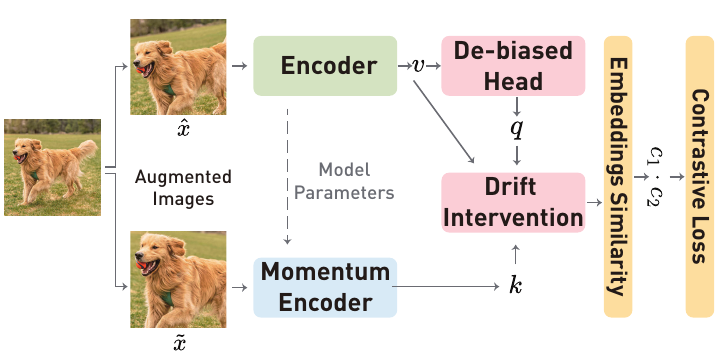}
        \caption{Our Resilient Contrastive Learning}
        \label{fig:mcl}
    \end{subfigure}    
    \caption{\textbf{Comparison between different contrastive pre-training paradigms.}
    (a) The key representations are sampled from a memory bank. (b) A momentum-updated encoder maintains the queue of keys.
    (c) The workflow of our resilient contrastive pre-training under concept drift streaming.
    Within the data streaming, a large batch size is opted for a wider drift adaptation window sliding to adapt changes in data distribution. Undergoes various random augmentations, the transformed instances from the identical sample are feature-extracted by both the encoder and the momentum encoder to get the key and value, respectively. A head is utilized to debias the drift and obtain the query of the encoder features. Subsequently, causal intervention is utilized to alleviate concept drift in the data stream within the adaptation window, resulting in the acquisition of two objects for contrastive learning.}
    \label{fig:workflow}
\end{figure*}

\subsection{Causal Interventional Contrastive Objective}

To eliminate the confounding influence of drift $D$ (the backdoor path) and recover the genuine causal effect $X \rightarrow Y$, we employ a causal intervention.
Drawing from the aforementioned causal relationships, it can be argued that current contrastive pertaining approaches involving the InfoNCE loss~\cite{oordRepresentationLearningContrastive2019} essentially leverages the likelihood $P(Y|X)$ to drive the whole network, which is vulnerable by the concept drift confounder $D$ and leads to the emergence of spurious correlations.  By marginalizing over the discrete drift contexts $D = \{d_1, \dots, d_{|D|}\}$~\cite{tangLongTailedClassificationKeeping2021}, $P_t(Y|X)$ can be expressed as:
\begin{equation}
    P_{t}(Y|X) = \sum_{i=1}^{|D|}P_{t}(Y|X, B=h(X,d_{i}))P_{t}(d_{i} | X)
\end{equation}
where $B=h(X,d)$ represents the sample bias $B$ conditioned on features $X$ and a specific drift context $d_i$. 
Following the~\cite{dengComprehensiveKnowledgeDistillation2021}, $d_i$ is simplified to the drift of the latent category center underlying the pre-training. The term $P_t(d|X)$ highlights how $X$ might carry information about $d$.


Inspired by recent advancements in applying causal inference to deep learning~\cite{lvCausalityInspiredRepresentation2022,choiC2LCausallyContrastive2022,rohekarCausalInterpretationSelfAttention2023}, we employ causal intervention $P_{t}(Y | \text{do}(X))$ to cut off the pathway of $D\rightarrow X$ at timestamp $t$ in drift data stream as illustrated in Fig.~\ref{fig:intervention}, where $\text{do}(\cdot)$ denotes the interventional operation~\cite{pearl2016causal, pearl2000models}.
In accordance with the backdoor adjustment, the de-confounding operation $\text{do}(X)$ involves stratifying the confounder into discrete segments $D = \{d_1, d_2, \ldots, d_{|D|}\}$, thereby rendering $D$ no longer a confounder between $X$ and $Y$. Thus, $P_{t}(Y | \text{do}(X))$ can be expressed as:
\begin{equation}
\label{eq.do}
    P_{t}(Y | \text{do}(X)) = \sum_{i=1}^{|D|}P_{t}(Y|X, B=h(X,d_{i}))P_{t}(d_{i})
\end{equation}

In Eq.~\eqref{eq.do}, $P_{t}(Y | \text{do}(X))$ compels $X$ to equally consider each confounding factor $d_{i}$ and integrates them collectively to predict $Y$. Then, the original classifier $P_{t}(Y|X)$ is replaced by $P_{t}(Y | \text{do}(X))$ at the iteration $t$, as shown:  
\begin{equation}
    P_{t}(Y | \text{do}(X)) \triangleq P_{t}(Y|X)
\end{equation}
The de-confound classifier mitigates the confounding influence and captures the genuine causality from $X$ to $Y$, thereby enhancing the quality of the contrastive pre-training. Nevertheless, Eq.~\eqref{eq.do} necessitates costly sampling to approximate $P(Y | \text{do}(X))$ when implementing it in contrastive pre-training, resulting in impractical training times. 
Fortunately, inspired by prior studies that approximate Eq.~\eqref{eq.do} in a feed-forward manner, we follow this line of work and employ a single-pass estimation to approximate the intervention effect:
\begin{equation}
\label{eq:inv}
    P_{t}(Y | \text{do}(X)) \approx P_{t}(Y | X, B =\sum_{i=1}^{|D|}  h(X,d_{i}) P_{t}(d_{i}))
\end{equation}
As a result, according to the interventional probability outlined in Eq.~\eqref{eq:inv}, contrastive pre-training is compelled to grasp the genuine causal effect: $X \rightarrow Y$ rather than the misleading correlations induced by the concept drift confounder $D$. Thus, combined with Eq. \eqref{eq:max}, the final interventional contrastive pre-training objective is formulated as:
\begin{equation}
\label{eq:obj}
\small
\begin{aligned}
     \mathcal{L}_{\text{agree}}(g_{j}^{\theta}(\tilde{X}_{j}), m_{j}^{\theta}(\hat{X}_{j})) = P_{t}(Y | X, \sum_{i=1}^{|D|}  h(X,d_{i}) P_{t}(d_{i})) \\
     = P_{t}(Y|g(\tilde{X}_{j}), \text{Softmax}(h(g_{j}^{\theta}(\tilde{X}_{j} m_{j}^{\theta}(\hat{X}_{j}) ))),
\end{aligned}     
\end{equation}
where the output of the final expression serves as the similarity score $\mathcal{L}_{\text{agree}}$ for the contrastive loss.

\subsection{Resilient Contrastive Pre-training}

This section details the concrete architecture and mechanism used to instantiate the causal objective (Eq. \eqref{eq:obj}).

\begin{table*}
    \caption{Evaluation results of fine-tuning on long-tailed classification tasks with ImageNet-LT~\cite{liuLargeScaleLongTailedRecognition2019}  and iNatualist2018~\cite{vanhornINaturalistSpeciesClassification2018a}. The best-performing contrastive pre-training results are highlighted in red. Many, Medium and Few denote the evaluated splits of many-shot ($>$100 training samples), medium-shot (20-100 samples) and few-shot ($<$20 samples). Top-1 accuracy is applied to evaluate the performance of different methods. $\dagger$  denotes methods that are adjusted for the long-tailed fine-tuning. }
    \centering
    \label{table:ft-lt}
    \setlength{\tabcolsep}{2mm}{
        \begin{tabular}{@{}lccccccccccc@{}}
        \toprule
        \multicolumn{1}{c}{}                                                                         &                             &                                                    & \multicolumn{4}{c}{ImageNet-LT}                                                                                       & \multicolumn{4}{c}{iNaturalist2018}                                                                                   \\
        \multicolumn{1}{c}{\multirow{-2}{*}{Methods}}                                                & \multirow{-2}{*}{Venue}     & \multirow{-2}{*}{Backbones}                        & Many                        & Medium                      & Few                         & All                         & Many                        & Medium                      & Few                         & All                         \\ \midrule
        \multicolumn{11}{l}{\textbf{{\color[HTML]{808080} Training from   scratch}}}                                                                                                                                                                                                                                                                                                                               \\ 
        {\color[HTML]{808080} cRT   \cite{kangDecouplingRepresentationClassifier2019}}               &  {\color[HTML]{808080} ICLR'19 }  & {\color[HTML]{808080} }                            & {\color[HTML]{808080} 61.8} & {\color[HTML]{808080} 46.2} & {\color[HTML]{808080} 27.3} & {\color[HTML]{808080} 49.6} & {\color[HTML]{808080} 69.0} & {\color[HTML]{808080} 66.0} & {\color[HTML]{808080} 63.2} & {\color[HTML]{808080} 65.2} \\
        {\color[HTML]{808080} LWS   \cite{kangDecouplingRepresentationClassifier2019}}               &  {\color[HTML]{808080} ICLR'19 }  & {\color[HTML]{808080} }                            & {\color[HTML]{808080} 60.2} & {\color[HTML]{808080} 47.2} & {\color[HTML]{808080} 30.3} & {\color[HTML]{808080} 49.9} & {\color[HTML]{808080} 65.0} & {\color[HTML]{808080} 66.3} & {\color[HTML]{808080} 65.5} & {\color[HTML]{808080} 65.9} \\
        {\color[HTML]{808080} RIDE   \cite{wangLongtailedRecognitionRouting2020}}                    &  {\color[HTML]{808080} ICLR'20 }  & {\color[HTML]{808080} }                            & {\color[HTML]{808080} 68.2} & {\color[HTML]{808080} 53.8} & {\color[HTML]{808080} 36.0} & {\color[HTML]{808080} 56.9} & {\color[HTML]{808080} 70.9} & {\color[HTML]{808080} 72.4} & {\color[HTML]{808080} 73.1} & {\color[HTML]{808080} 72.6} \\
        {\color[HTML]{808080} PaCo   \cite{cuiParametricContrastiveLearning2021}}                    &  {\color[HTML]{808080} ICCV'21 }  & \multirow{-4}{*}{{\color[HTML]{808080} ResNet-50}} & {\color[HTML]{808080} 68.2} & {\color[HTML]{808080} 58.7} & {\color[HTML]{808080} 41.0} & {\color[HTML]{808080} 60.0} & {\color[HTML]{808080} 70.3} & {\color[HTML]{808080} 73.2} & {\color[HTML]{808080} 73.6} & {\color[HTML]{808080} 73.2} \\ \midrule
        \multicolumn{11}{l}{\textbf{{\color[HTML]{808080} Masked Image   Modeling Pre-training}}}                                                                                                                                                                                                                                                                                                                                   \\ 
        {\color[HTML]{808080} MAE   \cite{heMaskedAutoencodersAre2022}}                              &  {\color[HTML]{808080} CVPR'22}   & {\color[HTML]{808080} }                            & {\color[HTML]{808080} 74.7} & {\color[HTML]{808080} 48.2} & {\color[HTML]{808080} 19.4} & {\color[HTML]{808080} 54.5} & {\color[HTML]{808080} 79.6} & {\color[HTML]{808080} 70.8} & {\color[HTML]{808080} 65.0} & {\color[HTML]{808080} 69.4} \\
        {\color[HTML]{808080} LiVT $\dagger$    \cite{xuLearningImbalancedData2023}}                 &  {\color[HTML]{808080} CVPR'23}   & \multirow{-2}{*}{{\color[HTML]{808080} ViT-B/16}}  & {\color[HTML]{808080} 73.6} & {\color[HTML]{808080} 56.4} & {\color[HTML]{808080} 41.0} & {\color[HTML]{808080} 60.9} & {\color[HTML]{808080} 78.9} & {\color[HTML]{808080} 76.5} & {\color[HTML]{808080} 74.8} & {\color[HTML]{808080} 76.1} \\ \midrule
        \multicolumn{11}{l}{\textbf{Contrastive   Learning Pre-training}}                                                                                                                                                                                                                                                                                                                                                           \\ 
        BYOL \cite{grillBootstrapYourOwn2020}                                                        &  NeurIPS'20                 &                                                    & 30.6                        & 8.6                         & 0.9                         & 15.9                        & 28.4                        & 22.6                        & 24.5                        & 24.0                        \\
        ViT   \cite{dosovitskiyImageWorth16x162021}                                                  &  NeurIPS'21                 &                                                    & 50.5                        & 23.5                        & 6.9                         & 31.6                        & 65.4                        & 55.3                        & 50.9                        & 54.6                        \\
        DINO   \cite{caronEmergingPropertiesSelfSupervised2021}                                      &  ICCV'21                    &                                                    & 64.8                        & 35.5                        & 11.3                        & 43.3                        & 76.7                        & 67.1                        & 61.2                        & 65.5                        \\
        DeiT \cite{touvronDeiTIIIRevenge2022}                                                        &  ECCV'22                    &                                                    & 70.4                        & 40.9                        & 12.8                        & 48.4                        & 72.9                        & 62.8                        & 55.8                        & 61.0                        \\
        MoCo v3   \cite{chenEmpiricalStudyTraining2021}                                              &  ICCV'21                    & \multirow{-4}{*}{ViT-B/16}                         & 70.8                        & 40.7                        & 14.3                        & 48.5                        & 76.6                        & 65.8                        & 62.8                        & 65.6                        \\
        DINO v2 \cite{zhouDINOWMWorldModels2024}                                                     &  ICML'25                    &                                                    & 68.4                        & 36.9                        & 11.7                        & 45.4                        & 71.9                        & 62.7                        & 58.2                        & 61.6                        \\
        \rowcolor[HTML]{D9D9D9}                                                                                                                                                                  
        Ours                                                                                         &                             & \multicolumn{1}{l}{\cellcolor[HTML]{D9D9D9}}       & {\color[HTML]{FE0000} 72.2} & {\color[HTML]{FE0000} 43.1} & {\color[HTML]{FE0000} 16.0} & {\color[HTML]{FE0000} 50.3} & {\color[HTML]{FE0000} 77.7} & {\color[HTML]{FE0000} 68.7} & {\color[HTML]{FE0000} 63.8} & {\color[HTML]{FE0000} 67.5} \\ \bottomrule
        \end{tabular}
    }
\end{table*}

\begin{table}
    \caption{Evaluation results of linear probing on long-tailed classification tasks with ImageNet-LT and iNatualist2018. The best-performing models are highlighted in red.}
    \centering
    \label{table:lb-lt}
    \setlength{\tabcolsep}{3mm}{
        \begin{tabular}{@{}lcccc@{}}
        \toprule
        Methods                                                 & Many                        & Medium                      & Few                         & All                         \\ \midrule
        \multicolumn{5}{l}{ImageNet-LT \cite{liuLargeScaleLongTailedRecognition2019}}                                                                                                                                                 \\ \midrule
        BYOL   \cite{grillBootstrapYourOwn2020}                 & 8.5                         & 0.5                         & 0.1                         & 3.5                         \\
        DINO   \cite{caronEmergingPropertiesSelfSupervised2021} & 47.4                        & 20.5                        & 0.1                         & 27.9                        \\
        DINO v2    \cite{zhouDINOWMWorldModels2024}             & 49.4                        & 23.4                        & 5.7                         & 30.8                        \\
        MoCo v3    \cite{chenEmpiricalStudyTraining2021}        & 45.8                        & 18.7                        & 0.0                         & 27.1                        \\
        ReLIC      \cite{mitrovicrepresentation}          & 3.2                         & 21.0                             & 47.1                        & 28.6 \\
        \rowcolor[HTML]{D9D9D9} 
        Ours                                                    & {\color[HTML]{FF0000} 52.7} & {\color[HTML]{FF0000} 24.0} & {\color[HTML]{FF0000} 7.9}  & {\color[HTML]{FF0000} 32.8} \\ \midrule
        \multicolumn{5}{l}{iNaturalist2018  \cite{vanhornINaturalistSpeciesClassification2018a}}                                                                                                                                             \\ \midrule
        BYOL   \cite{grillBootstrapYourOwn2020}                 & 7.0                         & 3.1                         & 2.5                         & 3.2                         \\
        DINO   \cite{caronEmergingPropertiesSelfSupervised2021} & 43.1                        & 33.5                        & 32.3                        & 34.0                        \\
        DINO v2    \cite{zhouDINOWMWorldModels2024}             & 36.3                        & 30.2                        & 30.0                        & 30.7                        \\
        MoCo v3    \cite{chenEmpiricalStudyTraining2021}        & 32.9                        & 25.4                        & 22.4                        & 24.9                        \\
        \rowcolor[HTML]{D9D9D9} 
        Ours                                                    & {\color[HTML]{FF0000} 48.4} & {\color[HTML]{FF0000} 37.0} & {\color[HTML]{FF0000} 35.1} & {\color[HTML]{FF0000} 37.3} \\ \bottomrule
        \end{tabular}
    }
\end{table}

To realize the causal interventional objective for CDP, we introduce \textbf{Resilient Contrastive Pre-training (RCP)}, an approach tailored for learning from data streams exhibiting concept drift. As depicted in Fig.~\ref{fig:workflow},
RCP adapts the student-teacher architecture of MoCo v3~\cite{chenEmpiricalStudyTraining2021}, utilizing an encoder ($g$) and a momentum encoder ($m$). Within each drift adaptation window $[t, t+\tau]$, two augmented views ($\tilde{x}$, $\hat{x}$) are generated per image.
The student encoder $g$ processes one view into features $v=g(\tilde{x})$, which an MLP head then projects into a query $q=h(v)$.
The momentum encoder $m$, structurally identical to $g$ and updated via EMA (Eq.~\eqref{eq:ema_cdp}), processes the other view to produce a key $k=m(\hat{x})$, ensuring stable targets and enhancing training stability.
Central to RCP is an intervention module that processes $q$, $v$, and $k$ to mitigate latent biases from concept drift. This module computes an interventional score component $c = \text{Softmax}(q \cdot k) \cdot v$, which embodies aspects of our causal objective (Eq.~\eqref{eq:obj}).
The overall network is then trained using an InfoNCE loss~\cite{oordRepresentationLearningContrastive2019}:
\begin{equation}
\label{eq:infoNCE}
    \mathcal{L} = -\log \frac{\exp{(c_{1}\cdot c_{2}^{i+}/\tau)}}{\sum_{i=t}^{t+\tau} \exp{(c_{1}\cdot c_{2}^{i}/\tau)}},
\end{equation}
where $\tau$ is a temperature hyper-parameter~\cite{wuUnsupervisedFeatureLearning2018a}. The sum is over one positive and all negative samples within the drift adaptation window $[t,t+\tau]$ of the data stream. 

Intuitively, $q$ akin to a "query"~\cite{heMomentumContrastUnsupervised2020}, the momentum output $k$ acts as the "key" for sampling key drift of the data stream, and the "value" $v$ resembles a blend of the essential contrastive object and the concept drift. Therefore, the goal of causal intervention is to mitigate the bias of the contrastive objective through the drift sampling within the adaption window. We find the derived causal intervention module has a similar structure to the self-attention mechanism~\cite{vaswaniAttentionAllYou2017}. Thus, this query-key-value mechanism structurally resembles self-attention~\cite{vaswaniAttentionAllYou2017}, suggesting an inherent capability to model contextual dependencies useful for tracking drift.  However, it is important to highlight that we employ "key" and "query" sampling for concept drift within the data stream to mitigate bias in the "value", which is different from the aim of sequence position modelling in the self-attention mechanism. 
Furthermore, our RCP method is simple to implement, efficient and scalable, with more details provided in Appendix.

\section{Experiments}

In this section, we initially showcase the robust performance of our RCP in the downstream task of long-tailed classification under drift pre-training, with results of fine-tuning and linear probing. Subsequently, from the perspective of generalization, we conduct experiments of out-of-distribution (OOD) detection and domain shift, to assess the resilience of the model against gradual drift and sudden drift. Following that, the feature embedding space of pre-trained models is intuitively visualized with degrees, illustrating how RCP mitigates the concept drift within the feature space. Finally, we present the scaling ability of our method facing with the drift pre-training. 

It is worth noting that we do not assume any type of drift. 
We chose several benchmarks with known drifts, such as long tail, OOD, and domain shift, only for comparison with existing benchmarks, and visually illustrated the effectiveness of our model. However, we did not optimize for specific drift types during training.
More detailed experimental implementations are given in Appendix.

\subsection{Navigating Tailed Drift Pre-training}


We compare our proposed RCP with other models to explicitly demonstrate its superior performance in tailed drift pre-training. Two large-scale datasets, ImageNet-LT~\cite{liuLargeScaleLongTailedRecognition2019} and iNaturalist 2018~\cite{vanhornINaturalistSpeciesClassification2018a}, are utilized as source datasets to perform pre-training, fine-tuning and linear probing, respectively. 
To effectively illustrate the efficacy of RCP in mitigating tail drift, we follow the criterion and metrics of long-tailed classification~\cite{yangTdistributedSphericalFeature2023} to report the Top-1 accuracy across various splits, including Many split with over 100 training samples, Medium split with 20-100 training samples, and Few split with fewer than 20 training samples.

In terms of fine-tuning results shown in Table~\ref{table:ft-lt}, we compare mainstream contrastive pre-training methods, namely DeiT~\cite{touvronDeiTIIIRevenge2022}, BYOL~\cite{grillBootstrapYourOwn2020} , DINO~\cite{caronEmergingPropertiesSelfSupervised2021} , DINO v2~\cite{zhouDINOWMWorldModels2024}, MoCo v3~\cite{chenEmpiricalStudyTraining2021}  with the baseline of ViT~\cite{dosovitskiyImageWorth16x162021}, and our RCP exhibits superior results beyond other contrastive pre-training methods. Compared to other methods that employ the student-teacher paradigm, such as BYOL, DINO, DINO v2, and MoCo v3, our RCP method demonstrates additional performance enhancements on both the Medium and Few splits, especially in the iNaturalist2018 dataset. It is demonstrated that the proposed causal interventional objective significantly alleviates the accumulated bias due to momentum updates of the teacher network. 
Meanwhile, under the concept drift scenario, BYOL without negative sample pairs has the inferior performance of long-tailed classification to the baseline ViT, while it is not noticeable in equilibrium~\cite{chenEmpiricalStudyTraining2021}. We argue that, in drift pre-training, the essence of negative sample pairs is similar to the causal interventional objective, taking into account the causal relationship of all drifts for each positive sample like de-confounding operation in Eq.~\eqref{eq.do} to delineate the boundary of the feature space. Relying solely on maximizing positive agreement as a pre-training strategy will cause the model to be overwhelmed by the head category, eventually exacerbating concept drift.

Besides, we also provide other fine-tuning results of long-tailed classification under scratch training and masked image modelling pre-training for comprehensive analysis, as shown in Table~\ref{table:ft-lt}. It is worth noting that our focus is on the impact of drift environments on contrastive pre-training. Therefore, we exclusively utilized RCP solely during the pre-training phase, without making any adjustments to the downstream classification tasks. Consequently, in comparison to other methods, all contrastive pre-training approaches exhibit relatively lower performance.
In the context of the masked image modeling strategy, it is found that contrastive learning is more susceptible to concept drift in the data stream. We attribute it to the fact that contrastive learning relies more heavily on global information within the data flow to construct feature representations. 

\begin{table*}
    \begin{minipage}[b]{0.34\textwidth}
    \caption{Evaluation results of domain shift on ImageNet-V2~\cite{rechtImageNetClassifiersGeneralize2019}, ImageNet-Sketch~\cite{wangLearningRobustGlobal2019} and ImageNet-R~\cite{hendrycksManyFacesRobustness2021}. The best-performing models are highlighted in red. Top-1/-5 accuracy are applied to evaluate the performance. }
    \label{table:ds}
    \centering
    \begin{adjustbox}{width=.99\textwidth}{
    \setlength{\tabcolsep}{0.4mm}{
    \begin{tabular}{@{}lccccc@{}}
    \toprule
    Methods & Many                        & Medium                      & Few                         & Top-1                       & Top-5                       \\ \midrule
    \multicolumn{6}{l}{\textbf{ImageNet-V2   \cite{rechtImageNetClassifiersGeneralize2019}}}                                                                               \\
    BYOL    \cite{grillBootstrapYourOwn2020}                 & 29.7                        & 7.8                         & 1.0                         & 15.2                        & 31.8                        \\
    DINO    \cite{caronEmergingPropertiesSelfSupervised2021} & 63.4                        & 33.6                        & 11.6                        & 41.9                        & 65.1                        \\
    DINO v2 \cite{zhouDINOWMWorldModels2024}                 & 66.6                        & 35.4                        & 11.9                        & 44.0                        & 66.9                        \\
    MoCo v3 \cite{chenEmpiricalStudyTraining2021}            & 62.9                        & 37.3                        & 12.6                        & 43.1                        & 66.0                        \\
    \rowcolor[HTML]{D9D9D9} 
    Ours    & {\color[HTML]{FF0000} 69.7} & {\color[HTML]{FF0000} 41.2} & {\color[HTML]{FF0000} 14.9} & {\color[HTML]{FF0000} 48.3} & {\color[HTML]{FF0000} 72.1} \\ \midrule
    \multicolumn{6}{l}{\textbf{ImageNet-Sketch~\cite{wangLearningRobustGlobal2019}}}                                                                                      \\
    BYOL    \cite{grillBootstrapYourOwn2020}                 & 1.9                         & 0.2                         & 0.0                         & 0.8                         & 2.8                         \\
    DINO    \cite{caronEmergingPropertiesSelfSupervised2021} & 12.4                        & 5.0                         & 0.5                         & 7.1                         & 16.2                        \\
    DINO v2 \cite{zhouDINOWMWorldModels2024}                 & 17.1                        & 6.8                         & 1.4                         & 9.9                         & 20.8                        \\
    MoCo v3 \cite{chenEmpiricalStudyTraining2021}            & 17.6                        & 7.6                         & 1.4                         & 10.4                        & 21.9                        \\
    \rowcolor[HTML]{D9D9D9} 
    Ours    & {\color[HTML]{FF0000} 22.2} & {\color[HTML]{FF0000} 9.7}  & {\color[HTML]{FF0000} 2.3}  & {\color[HTML]{FF0000} 13.4} & {\color[HTML]{FF0000} 27.0} \\ \midrule
    \multicolumn{6}{l}{\textbf{ImageNet-R~\cite{hendrycksManyFacesRobustness2021}}}                                                                                      \\
    BYOL    \cite{grillBootstrapYourOwn2020}                 & 4.2                         & 1.7                         & 0.0                         & 2.7                         & 6.8                         \\
    DINO    \cite{caronEmergingPropertiesSelfSupervised2021} & 13.4                        & 7.6                         & 0.4                         & 9.5                         & 18.3                        \\
    DINO v2 \cite{zhouDINOWMWorldModels2024}                 & 15.8                        & 8.6                         & 0.7                         & 11.1                        & 21.1                        \\
    MoCo v3 \cite{chenEmpiricalStudyTraining2021}            & 17.2                        & 9.8                         & 1.0                         & 12.3                        & 22.8                        \\
    \rowcolor[HTML]{D9D9D9} 
    Ours    & {\color[HTML]{FF0000} 21.3} & {\color[HTML]{FF0000} 12.5} & {\color[HTML]{FF0000} 1.6}  & {\color[HTML]{FF0000} 15.5} & {\color[HTML]{FF0000} 27.5} \\ \bottomrule
    \end{tabular}
    }        
    }\end{adjustbox}
    \end{minipage}
    \hfill
    \begin{minipage}[b]{0.65\textwidth}
    \caption{Evaluation results of OOD detection with the OOD datasets of Texture~\cite{cimpoi2014describing}, iNat-OOD~\cite{van2018inaturalist,huangMOSScalingOutofDistribution2021}, ImageNet-O~\cite{hendrycksNaturalAdversarialExamples2021} and OpenImage-O~\cite{wangViMOutofDistributionVirtualLogit2022}. ViT-B/16 is selected as the image encoder. The best-performing method is highlighted in red. FPR$\downarrow$ and AUROC$\uparrow$ are applied to evaluate the performance of different methods. PT denotes the pre-training dataset, while ID means the in-distribution dataset.}
    \centering
    \label{table:ood}
    \begin{adjustbox}{width=.99\textwidth}{
    \setlength{\tabcolsep}{0.5mm}{
    \begin{tabular}{@{}lcccccccccc@{}}
    \toprule
                                                            & \multicolumn{2}{c}{Texture}                               & \multicolumn{2}{c}{iNaturalist}                           & \multicolumn{2}{c}{ImageNet-O}                            & \multicolumn{2}{c}{OpenImage-O}                           & \multicolumn{2}{c}{Overall}                               \\
    \multirow{-2}{*}{Methods}                               & AUROC                       & FPR                         & AUROC                       & FPR                         & AUROC                       & FPR                         & AUROC                       & FPR                         & AUROC                       & FPR                         \\ \midrule
    \multicolumn{11}{l}{\textbf{PT: ImageNet-21k, ID: ImageNet-1K}}  \\
    MSP   \cite{hendrycks2016baseline}                      & 71.3                        & 77.1                        & 90.7                        & 43.7                        & 60.8                        & 90.6                        & 84.3                        & 61.8                        & 76.8                        & 68.3                        \\
    Energy   \cite{liu2020energy}                           & 54.1                        & 86.3                        & 76.6                        & 72.7                        & 61.6                        & 81.0                        & 71.1                        & 74.0                        & 65.9                        & 78.5                        \\
    MaxLog   \cite{hendrycks2019scaling}                    & 67.2                        & 78.0                        & 89.9                        & 45.6                        & 61.7                        & 88.6                        & 82.7                        & 62.5                        & 75.4                        & 68.7                        \\
    KL   \cite{hendrycks2019scaling}                        & 82.6                        & 67.3                        & 87.6                        & 69.7                        & 66.6                        & 88.2                        & 84.3                        & 74.2                        & 80.3                        & 74.8                        \\
    Residual \cite{wang2022vim}                             & 82.4                        & 64.6                        & 73.7                        & 86.0                        & 68.4                        & 87.5                        & 74.9                        & 78.0                        & 74.9                        & 79.0                        \\
    React \cite{sun2021react}                               & 62.1                        & 80.5                        & 91.2                        & 38.7                        & 63.7                        & 81.0                        & 80.4                        & 60.4                        & 74.3                        & 65.2                        \\
    Mahalanobis  \cite{lee2018simple}                       & 84.9                        & 66.1                        & 84.9                        & 81.6                        & 71.5                        & 88.9                        & 84.2                        & 74.7                        & 81.4                        & 77.8                        \\
    ViM \cite{wang2022vim}                                  & 83.5                        & 62.7                        & 77.8                        & 81.7                        & 71.0                        & 86.6                        & 78.3                        & 74.6                        & 77.7                        & 76.4                        \\
    MOODv1 \cite{li2023rethinking}                          & 93.0                        & 30.9                        & 98.8                        & 5.9                         & 86.8                        & 63.2                        & 95.5                        & 26.5                        & 93.5                        & 31.6                        \\
    MOODv2  \cite{liMOODv2MaskedImage2024}                  & 94.3                        & 24.7                        & 99.6                        & 1.8                         & 91.5                        & 40.8                        & 97.4                        & 13.6                        & 95.7                        & 20.2                        \\ \midrule
    \multicolumn{11}{l}{\textbf{PT: ImageNet-LT, ID: ImageNet-LT}}  \\
    BYOL   \cite{grillBootstrapYourOwn2020}                 & 58.4                        & 92.1                        & 80.1                        & 77.5                        & 52.0                        & 92.2                        & 65.3                        & 81.3                        & 64.0                        & 85.8                        \\
    DINO   \cite{caronEmergingPropertiesSelfSupervised2021} & 72.9                        & 80.7                        & 85.7                        & {\color[HTML]{FF0000} 65.3} & {\color[HTML]{FF0000} 74.2} & {\color[HTML]{FF0000} 80.2} & 74.7                        & 83.0                        & 76.9                        & 77.3                        \\
    DINO v2    \cite{zhouDINOWMWorldModels2024}             & 67.1                        & 85.1                        & 78.9                        & 88.4                        & 63.1                        & 88.3                        & 73.9                        & 85.4                        & 70.7                        & 86.8                        \\
    MoCo v3    \cite{chenEmpiricalStudyTraining2021}        & 81.5                        & 63.9                        & 80.1                        & 68.6                        & 65.0                        & 91.0                        & 69.9                        & 84.2                        & 74.1                        & 76.9                        \\
    \rowcolor[HTML]{D9D9D9} 
    Ours                                                    & {\color[HTML]{FF0000} 83.1} & {\color[HTML]{FF0000} 57.8} & {\color[HTML]{FF0000} 88.0} & 67.7                        & 70.1                        & 82.9                        & {\color[HTML]{FF0000} 75.5} & {\color[HTML]{FF0000} 78.3} & {\color[HTML]{FF0000} 79.2} & {\color[HTML]{FF0000} 71.7} \\ \bottomrule
    \end{tabular}
    }
    }\end{adjustbox}
    \end{minipage}
\end{table*}

Furthermore, we provide linear probing results of contrastive pre-training methods in Table~\ref{table:lb-lt}. It is evidenced that our resilient contrastive learning method outperforms other contrastive pre-training methods, corroborating our main contributions of causal interventional contrastive objective for drift pre-training. It is noteworthy that on the Few split of ImageNet-LT, several contrastive learning methods failed to perform effectively, whereas we attained an accuracy of 7.9\%. It indicates that our approach can accurately construct feature representations for tail categories even in the presence of concept drift.

\subsection{Taming ID and OOD Drift}

To validate the generalization capability of RCP under drifting scenarios, we conducted experiments from the perspectives of domain shift and out-of-distribution (OOD) detection.

In terms of domain shift generalization as exhibited in Table~\ref{table:ds}, our primary focus is to assess the extent of the drift environment on the feature representation constructed during pre-training. Therefore, we select three subsets of ImageNet~\cite{russakovskyImagenetLargeScale2015} to validate the domain shift performance, namely, ImageNet-V2~\cite{rechtImageNetClassifiersGeneralize2019}, ImageNet-Sketch~\cite{wangLearningRobustGlobal2019} and ImageNet-R~\cite{hendrycksManyFacesRobustness2021}. Our results achieve superior results across three datasets, investigating the effectiveness of the proposed RCP method. It corroborates that, the feature space we construct in pre-training abstracts and captures the essential information of images, while disregarding any unexpected interference from the concept drift within the data stream.
Moreover, judging from the results of Few split in ImageNet-Sketch and ImageNet-R, the majority of contrastive pre-training methods struggle to differentiate the tail categories. This difficulty arises from two main challenges. First, the bias induced by tail drift tends to skew models toward favoring head categories. On the flip side, the small sample size of tail categories limits the ability of pre-training methods to accurately extract their fundamental features, a problem that is exacerbated in the presence of domain shifts. Accordingly, our outstanding results demonstrate our proficiency in mitigating tail drift and extracting crucial features from limited samples.

Concerning out-of-distribution (OOD) detection, we also evaluate the capability of our RCP approach to delineate the boundaries of the feature space, as presented in Table \ref{table:ood}. While exhibiting slightly lower performance than DINO on ImageNet-O, we significantly outperform contrastive pre-training methods on the other three out-of-distribution datasets, and surpasses other approaches in overall performance. It demonstrates that our model possesses the benefits of intra-class compactness and inter-class separability in feature representation. 
Meanwhile, it is worth noting that in larger-scale balanced pre-training datasets, such as ImageNet-21k~\cite{ridnikImageNet21KPretrainingMasses2021}, our approach outperforms numerous traditional OOD detection methods in overall performance, such as ViM~\cite{wang2022vim}. It underscores our RCP method can effectively alleviate concept drift bias in the model while also enhancing the characterization of the model's decision boundary.


\begin{table*}[htbp]
\caption{Evaluation results of different contrastive learning strategies in the stage of pre-training from three perspectives: ID intra-class compactness, ID inter-class separability and the separability between ID and OOD categories.     
The cosine metric is utilized to measure these distances, which is expressed as average degrees. ImageNet-LT is utilized for pre-training. Best results are highlighted in red.}
\label{table:pt}
\centering
\begin{tabular}{@{}lcccccccccccc@{}}
\toprule
\multicolumn{1}{c}{}                                    & \multicolumn{4}{c}{ID Intra Compactness $\downarrow$}                                                         & \multicolumn{4}{c}{ID Inter   Separability $\uparrow$}                                                          & \multicolumn{4}{c}{ID vs. OOD   Separability $\uparrow$}                                                              \\
\multicolumn{1}{c}{\multirow{-2}{*}{Pre-training}}      & Many                        & Medium                      & Few                         & All                         & Many                        & Medium                      & Few                         & All                         & Many                        & Medium                      & Few                         & All                         \\ \midrule
BYOL \cite{grillBootstrapYourOwn2020}                   & 53.1                        & 59.1                        & 59.9                        & 56.9                        & 81.5                        & 79.2                        & 78.3                        & 80.0                        & 75.7                        & 71.9                        & 70.3                        & 73.1                        \\
DINO   \cite{caronEmergingPropertiesSelfSupervised2021} & 34.2                        & 51.4                        & 59.0                        & 45.9                        & 89.0                        & 88.4                        & 87.6                        & 88.5                        & 86.2                        & 84.4                        & 81.4                        & 84.7                        \\
DINO v2 \cite{zhouDINOWMWorldModels2024}                & 31.2                        & 50.1                        & 57.8                        & 43.9                        & 88.9                        & 88.1                        & 87.2                        & 88.3                        & 86.0                        & 83.7                        & 80.5                        & 84.1                        \\
MoCo v3   \cite{chenEmpiricalStudyTraining2021}         & 29.1                        & 47.8                        & 57.8                        & 42.8                        & 89.3                        & 88.8                        & 88.2                        & 88.9                        & 86.9                        & 85.7                        & 83.0                        & 85.8                        \\
\rowcolor[HTML]{D9D9D9} 
Ours                                                    & {\color[HTML]{FF0000} 28.5} & {\color[HTML]{FF0000} 46.5} & {\color[HTML]{FF0000} 55.8} & {\color[HTML]{FF0000} 40.9} & {\color[HTML]{FF0000} 89.4} & {\color[HTML]{FF0000} 89.0} & {\color[HTML]{FF0000} 88.4} & {\color[HTML]{FF0000} 89.1} & {\color[HTML]{FF0000} 87.1} & {\color[HTML]{FF0000} 86.1} & {\color[HTML]{FF0000} 83.4} & {\color[HTML]{FF0000} 86.1} \\ \bottomrule
\end{tabular}
\end{table*}

\subsection{Boosting Pre-training Feature Embedding}
To directly and intuitively demonstrate the feature space of the pre-trained model, we quantitatively calculate the feature embedding distances from three perspectives: In-Distribution (ID) intra-class compactness, ID inter-class separability, and the separability between ID and OOD categories, as shown in Table~\ref{table:pt}.
Intra-class compactness within the In-Distribution (ID) measures the average distance between the category center and samples. ID Inter-class separability evaluates the distances between centers of different categories. Lastly, we assess the separability between ID categories and OOD samples.

The intra-class compactness results demonstrate that our model effectively validates the efficacy of the proposed RCP in delineating feature boundaries, especially in long-tailed scenarios. Moreover, the inter-class separability among different splits is very similar, suggesting that the pre-training primarily influences intra-class compactness rather than inter-class separability.
Furthermore, with the exception of BYOL, the inter-class separability of the other contrastive pre-training methods is nearly the same. This observation underscores the significance of negative samples in expanding inter-class separation.
It also demonstrates that the bias is induced by accumulated drift within momentum updates, where features of head categories dominated the whole model under tailed drift. In the context of ID vs. OOD separability, we attained the optimal results, signifying that our feature space exhibits distinct feature boundaries. 
In addition, we find that the performance of ID vs. OOD separability aligns with the ID intra-class compactness, suggesting that enhancing the model's performance hinges on mitigating data stream drift to enhance feature extraction capabilities.

\subsection{Scaling with Drift Pre-training}

\begin{table}[htb]
\caption{Evaluation results of scaling ability of our RCP on ImageNet-LT with various ViT models. Top-1 accuracy is utilized to evaluate the performance.}
\label{table:scale}
\centering
\setlength{\tabcolsep}{2mm}{
\begin{tabular}{@{}lcccc@{}}
\toprule
Backbones                          & Many                 & Medium               & Few                  & All                  \\ \midrule
ViT-Small/16  & 67.7 & 39.6 & 11.6 & 46.4 \\
ViT-Base/16  & 72.2 & 43.1 & 16.0 & 50.3 \\
ViT-Large/16  & 73.8 & 45.0 & 16.9 & 52.0 \\ \bottomrule
\end{tabular}
}
\end{table}

To present the scalability of our RCP approach in scenarios involving concept drift, we evaluate the efficiency and scalability of our model in Table~\ref{table:scale}. As the number of parameters increases, our model exhibits a scalable effect, leading to performance improvements with larger models in downstream classification tasks. In comparison to ViT-S/16, our proposed method achieves nearly a 4\% enhancement with ViT-B/16. This suggests that we can effectively train larger models to attain superior performance in the face of concept drift within data streams.

More critically, our experiments reveal that pre-trained models exhibit consistent scaling behavior even on datasets with great distribution drift. It demonstrates that our proposed RCP can effectively leverage larger-scale, minimally curated datasets, thereby substantially expanding the pool of usable data for pre-training without relying on complicated cleaning procedures.

\subsection{Ablation on Adaptation Window Length}

\begin{table}[htbp]
\centering
\caption{Ablation experiments on adaptation window length within the linear probing task on ImageNet-LT. We also provide the results of MoCo v3 for comparison.}
\label{tab:ab}
\begin{tabular}{@{}lccccc@{}}
\toprule
        & Window  & Many & Medium & Few & All  \\ \midrule
MoCo v3 \cite{chenEmpiricalStudyTraining2021} & 4,096         & 45.8 & 18.7   & 0.0 & 27.1 \\
Ours    & 2,048         & 42.1 & 12.8   & 1.1 & 22.4 \\
Ours    & 4,096         & 47.9 & 20.1   & 3.2 & 28.5 \\
Ours    & 9,600         & 52.7 & 24.0   & 7.9 & 32.8 \\ \bottomrule
\end{tabular}
\end{table}

Furthermore, we also conduct ablation experiments on adaptation window length, as exhibited in Table \ref{tab:ab}. It demonstrates that our approach relies on a wider drift adaptation window, with superior results under a larger one.  In particular, our method consistently outperforms MoCo v3 on Few Split, demonstrating the robustness of our method against non-stationary drift.

\section{Conclusion and Outlooks}

In this paper, we present resilient contrastive pre-training (RCP), a novel, straightforward and effective pre-training paradigm tailored for concept drift data stream. We employ causal inference theory to methodically examine the source of bias in the momentum update of contrastive pre-training and put forward a causal interventional objective to mitigate this bias within the drifting data stream. By virtue of this objective, RCP is devised to counteract the unpredictable distribution changes occurring within the data stream.


{
    \small
    \bibliographystyle{ieeenat_fullname}
    \bibliography{main}

@String(ECCV= {Eur. Conf. Comput. Vis.})

@String(ICLR = {Int. Conf. Learn. Represent.})

@String(AAAI = {AAAI})

@String(ECCV  = {ECCV})

@String(ICLR  = {ICLR})

@inproceedings{caronEmergingPropertiesSelfSupervised2021,
  title = {Emerging {{Properties}} in {{Self-Supervised Vision Transformers}}},
  author = {Caron, Mathilde and Touvron, Hugo and Misra, Ishan and Jégou, Hervé and Mairal, Julien and Bojanowski, Piotr and Joulin, Armand},
  year = {2021},
  pages = {9650--9660},
  url = {https://openaccess.thecvf.com/content/ICCV2021/html/Caron_Emerging_Properties_in_Self-Supervised_Vision_Transformers_ICCV_2021_paper},
  urldate = {2024-09-03},
  booktitle = {Proceedings of the {{IEEE}}/{{CVF International Conference}} on {{Computer Vision}}},
  langid = {english}
}

@online{zhouDINOWMWorldModels2024,
  title = {{{DINO-WM}}: {{World Models}} on {{Pre-trained Visual Features}} Enable {{Zero-shot Planning}}},
  shorttitle = {{{DINO-WM}}},
  author = {Zhou, Gaoyue and Pan, Hengkai and LeCun, Yann and Pinto, Lerrel},
  year = {2024},
  eprint = {2411.04983},
  eprinttype = {arXiv},
  doi = {10.48550/arXiv.2411.04983},
  url = {http://arxiv.org/abs/2411.04983},
  urldate = {2024-12-14},
  langid = {american},
  pubstate = {prepublished}
}

@inproceedings{chenEmpiricalStudyTraining2021,
  title = {An {{Empirical Study}} of {{Training Self-Supervised Vision Transformers}}},
  author = {Chen, Xinlei and Xie, Saining and He, Kaiming},
  year = {2021},
  pages = {9640--9649},
  url = {https://openaccess.thecvf.com/content/ICCV2021/html/Chen_An_Empirical_Study_of_Training_Self-Supervised_Vision_Transformers_ICCV_2021_paper.html},
  urldate = {2024-09-03},
  booktitle = {Proceedings of the {{IEEE}}/{{CVF International Conference}} on {{Computer Vision}}},
  langid = {english}
}

@inproceedings{heMomentumContrastUnsupervised2020,
  title = {Momentum {{Contrast}} for {{Unsupervised Visual Representation Learning}}},
  author = {He, Kaiming and Fan, Haoqi and Wu, Yuxin and Xie, Saining and Girshick, Ross},
  year = {2020},
  pages = {9729--9738},
  url = {https://openaccess.thecvf.com/content_CVPR_2020/html/He_Momentum_Contrast_for_Unsupervised_Visual_Representation_Learning_CVPR_2020_paper.html},
  urldate = {2024-12-05},
  booktitle = {Proceedings of the {{IEEE}}/{{CVF Conference}} on {{Computer Vision}} and {{Pattern Recognition}}},
  langid = {american}
}

@inproceedings{chenSimpleFrameworkContrastive2020,
  title = {A {{Simple Framework}} for {{Contrastive Learning}} of {{Visual Representations}}},
  booktitle = {Proceedings of the 37th {{International Conference}} on {{Machine Learning}}},
  author = {Chen, Ting and Kornblith, Simon and Norouzi, Mohammad and Hinton, Geoffrey},
  year = {2020},
  pages = {1597--1607},
  publisher = {PMLR},
  url = {https://proceedings.mlr.press/v119/chen20j.html},
  urldate = {2024-08-30},
  eventtitle = {International {{Conference}} on {{Machine Learning}}},
  langid = {english}
}

@book{pearl2016causal,
  title={Causal inference in statistics: a primer},
  author={Pearl, Judea},
  year={2016},
  publisher={John Wiley \& Sons}
}

@incollection{pearl2022direct,
  title={Direct and indirect effects},
  author={Pearl, Judea},
  booktitle={Probabilistic and causal inference: the works of Judea Pearl},
  pages={373--392},
  year={2022}
}

@article{pearl1995causal,
  title={Causal diagrams for empirical research},
  author={Pearl, Judea},
  journal={Biometrika},
  volume={82},
  number={4},
  pages={669--688},
  year={1995},
  publisher={Oxford University Press}
}

@inproceedings{liuLargeScaleLongTailedRecognition2019,
  title = {Large-{{Scale Long-Tailed Recognition}} in an {{Open World}}},
  booktitle = {Proceedings of the {{IEEE}}/{{CVF Conference}} on {{Computer Vision}} and {{Pattern Recognition}}},
  author = {Liu, Ziwei and Miao, Zhongqi and Zhan, Xiaohang and Wang, Jiayun and Gong, Boqing and Yu, Stella X.},
  year = {2019},
  pages = {2532--2541},
  publisher = {IEEE},
  location = {Long Beach, CA, USA},
  doi = {10.1109/CVPR.2019.00264},
  url = {https://ieeexplore.ieee.org/document/8953407/},
  urldate = {2021-06-02},
  eventtitle = {Proceedings of the {{IEEE}}/{{CVF Conference}} on {{Computer Vision}} and {{Pattern Recognition}}},
  isbn = {978-1-7281-3293-8},
  langid = {english}
}

@article{russakovskyImagenetLargeScale2015,
  title = {Imagenet Large Scale Visual Recognition Challenge},
  author = {Russakovsky, Olga and Deng, Jia and Su, Hao and Krause, Jonathan and Satheesh, Sanjeev and Ma, Sean and Huang, Zhiheng and Karpathy, Andrej and Khosla, Aditya and Bernstein, Michael},
  year = {2015},
  journaltitle = {International journal of computer vision},
  volume = {115},
  number = {3},
  pages = {211--252}
}

@article{pearl2014interpretation,
  title={Interpretation and identification of causal mediation.},
  author={Pearl, Judea},
  journal={Psychological methods},
  volume={19},
  number={4},
  pages={459},
  year={2014},
  publisher={American Psychological Association}
}

@article{luLearningConceptDrift2019,
  title = {Learning under {{Concept Drift}}: {{A Review}}},
  shorttitle = {Learning under {{Concept Drift}}},
  author = {Lu, Jie and Liu, Anjin and Dong, Fan and Gu, Feng and Gama, João and Zhang, Guangquan},
  year = {2019},
  journaltitle = {IEEE Transactions on Knowledge and Data Engineering},
  volume = {31},
  number = {12},
  pages = {2346--2363},
  issn = {1558-2191},
  doi = {10.1109/TKDE.2018.2876857},
  url = {https://ieeexplore.ieee.org/abstract/document/8496795},
  urldate = {2024-02-21}
}

@article{pearl2000models,
  title={Models, reasoning and inference},
  author={Pearl, Judea and others},
  journal={Cambridge, UK: CambridgeUniversityPress},
  volume={19},
  number={2},
  pages={3},
  year={2000}
}

@article{oordRepresentationLearningContrastive2019,
  title={Representation learning with contrastive predictive coding},
  author={Oord, Aaron van den and Li, Yazhe and Vinyals, Oriol},
  journal={arXiv preprint arXiv:1807.03748},
  year={2018}
}

@inproceedings{yangCausalAttentionVisionLanguage2021,
  title = {Causal {{Attention}} for {{Vision-Language Tasks}}},
  author = {Yang, Xu and Zhang, Hanwang and Qi, Guojun and Cai, Jianfei},
  year = {2021},
  pages = {9847--9857},
  url = {https://openaccess.thecvf.com/content/CVPR2021/html/Yang_Causal_Attention_for_Vision-Language_Tasks_CVPR_2021_paper.html},
  urldate = {2024-12-31},
  booktitle = {Proceedings of the {{IEEE}}/{{CVF Conference}} on {{Computer Vision}} and {{Pattern Recognition}}},
  langid = {english}
}

@article{rohekarCausalInterpretationSelfAttention2023,
  title = {Causal {{Interpretation}} of {{Self-Attention}} in {{Pre-Trained Transformers}}},
  author = {Rohekar, Raanan Y. and Gurwicz, Yaniv and Nisimov, Shami},
  year = {2023},
  journaltitle = {Advances in Neural Information Processing Systems},
  volume = {36},
  pages = {31450--31465},
  url = {https://proceedings.neurips.cc/paper_files/paper/2023/hash/642a321fba8a0f03765318e629cb93ea\\-Abstract-Conference.html},
  urldate = {2024-12-30},
  langid = {english}
}

@inproceedings{lvCausalityInspiredRepresentation2022,
  title = {Causality {{Inspired Representation Learning}} for {{Domain Generalization}}},
  author = {Lv, Fangrui and Liang, Jian and Li, Shuang and Zang, Bin and Liu, Chi Harold and Wang, Ziteng and Liu, Di},
  year = {2022},
  pages = {8046--8056},
  url = {https://openaccess.thecvf.com/content/CVPR2022/html/Lv_Causality_Inspired_Representation_Learning_for_Domain_Generalization_CVPR_2022_paper.html},
  urldate = {2024-12-30},
  booktitle = {Proceedings of the {{IEEE}}/{{CVF Conference}} on {{Computer Vision}} and {{Pattern Recognition}}},
  langid = {english}
}

@online{miaoDomainGeneralizationContrastive2022,
  title = {Domain {{Generalization}} via {{Contrastive Causal Learning}}},
  author = {Miao, Qiaowei and Yuan, Junkun and Kuang, Kun},
  year = {2022},
  eprint = {2210.02655},
  eprinttype = {arXiv},
  eprintclass = {cs},
  doi = {10.48550/arXiv.2210.02655},
  url = {http://arxiv.org/abs/2210.02655},
  urldate = {2024-12-30},
  pubstate = {prepublished}
}

@article{choiC2LCausallyContrastive2022,
  title = {{{C2L}}: {{Causally Contrastive Learning}} for {{Robust Text Classification}}},
  shorttitle = {{{C2L}}},
  author = {Choi, Seungtaek and Jeong, Myeongho and Han, Hojae and Hwang, Seung-won},
  year = {2022},
  journaltitle = {Proceedings of the AAAI Conference on Artificial Intelligence},
  volume = {36},
  number = {10},
  pages = {10526--10534},
  issn = {2374-3468},
  doi = {10.1609/aaai.v36i10.21296},
  url = {https://ojs.aaai.org/index.php/AAAI/article/view/21296},
  urldate = {2024-12-30},
  issue = {10},
  langid = {english}
}

@inproceedings{liuShowDeconfoundTell2022,
  title = {Show, {{Deconfound}} and {{Tell}}: {{Image Captioning With Causal Inference}}},
  shorttitle = {Show, {{Deconfound}} and {{Tell}}},
  author = {Liu, Bing and Wang, Dong and Yang, Xu and Zhou, Yong and Yao, Rui and Shao, Zhiwen and Zhao, Jiaqi},
  year = {2022},
  pages = {18041--18050},
  url = {https://openaccess.thecvf.com/content/CVPR2022/html/Liu_Show_Deconfound_and_Tell_Image_Captioning_With_Causal_Inference_CVPR_2022_paper.html},
  urldate = {2024-12-12},
  booktitle = {Proceedings of the {{IEEE}}/{{CVF Conference}} on {{Computer Vision}} and {{Pattern Recognition}}},
  langid = {english}
}

@article{yangDeconfoundedImageCaptioning2023,
  title = {Deconfounded {{Image Captioning}}: {{A Causal Retrospect}}},
  shorttitle = {Deconfounded {{Image Captioning}}},
  author = {Yang, Xu and Zhang, Hanwang and Cai, Jianfei},
  year = {2023},
  journaltitle = {IEEE Transactions on Pattern Analysis and Machine Intelligence},
  volume = {45},
  number = {11},
  pages = {12996--13010},
  issn = {1939-3539},
  doi = {10.1109/TPAMI.2021.3121705},
  url = {https://ieeexplore.ieee.org/abstract/document/9583890},
  urldate = {2025-01-09},
  eventtitle = {{{IEEE Transactions}} on {{Pattern Analysis}} and {{Machine Intelligence}}}
}

@inproceedings{wuUnsupervisedFeatureLearning2018a,
  title = {Unsupervised {{Feature Learning}} via {{Non-Parametric Instance Discrimination}}},
  author = {Wu, Zhirong and Xiong, Yuanjun and Yu, Stella X. and Lin, Dahua},
  year = {2018},
  pages = {3733--3742},
  url = {https://openaccess.thecvf.com/content_cvpr_2018/html/Wu_Unsupervised_Feature_Learning_CVPR_2018_paper.html},
  urldate = {2025-01-11},
  booktitle = {Proceedings of the {{IEEE Conference}} on {{Computer Vision}} and {{Pattern Recognition}}}
}

@online{dosovitskiyImageWorth16x162021,
  title = {An {{Image}} Is {{Worth}} 16x16 {{Words}}: {{Transformers}} for {{Image Recognition}} at {{Scale}}},
  shorttitle = {An {{Image}} Is {{Worth}} 16x16 {{Words}}},
  author = {Dosovitskiy, Alexey and Beyer, Lucas and Kolesnikov, Alexander and Weissenborn, Dirk and Zhai, Xiaohua and Unterthiner, Thomas and Dehghani, Mostafa and Minderer, Matthias and Heigold, Georg and Gelly, Sylvain and Uszkoreit, Jakob and Houlsby, Neil},
  year = {2021},
  eprint = {2010.11929},
  eprinttype = {arxiv},
  doi = {10.48550/arXiv.2010.11929},
  url = {http://arxiv.org/abs/2010.11929},
  urldate = {2024-03-11},
  pubstate = {preprint}
}

@inproceedings{wangLongtailedRecognitionRouting2020,
  title = {Long-Tailed {{Recognition}} by {{Routing Diverse Distribution-Aware Experts}}},
  booktitle = {International {{Conference}} on {{Learning Representations}}},
  author = {Wang, Xudong and Lian, Long and Miao, Zhongqi and Liu, Ziwei and Yu, Stella},
  year = {2020},
  url = {https://openreview.net/forum?id=D9I3drBz4UC},
  urldate = {2021-11-03},
  eventtitle = {International {{Conference}} on {{Learning Representations}}},
  langid = {english}
}

@inproceedings{cuiParametricContrastiveLearning2021,
  title = {Parametric {{Contrastive Learning}}},
  author = {Cui, Jiequan and Zhong, Zhisheng and Liu, Shu and Yu, Bei and Jia, Jiaya},
  year = {2021},
  pages = {715--724},
  url = {https://openaccess.thecvf.com/content/ICCV2021/html/Cui_Parametric_Contrastive_Learning_ICCV_2021_paper.html},
  urldate = {2022-06-08},
  booktitle = {Proceedings of the {{IEEE}}/{{CVF International Conference}} on {{Computer Vision}}},
  langid = {english}
}

@inproceedings{kangDecouplingRepresentationClassifier2019,
  title = {Decoupling {{Representation}} and {{Classifier}} for {{Long-Tailed Recognition}}},
  booktitle = {Eighth {{International Conference}} on {{Learning Representations}} ({{ICLR}})},
  author = {Kang, Bingyi and Xie, Saining and Rohrbach, Marcus and Yan, Zhicheng and Gordo, Albert and Feng, Jiashi and Kalantidis, Yannis},
  year = {2019},
  url = {https://openreview.net/forum?id=r1gRTCVFvB},
  urldate = {2021-11-03},
  eventtitle = {International {{Conference}} on {{Learning Representations}}},
  langid = {english}
}

@inproceedings{yu2024online,
  title={Online Boosting Adaptive Learning under Concept Drift for Multistream Classification},
  author={Yu, En and Lu, Jie and Zhang, Bin and Zhang, Guangquan},
  booktitle={Proceedings of the AAAI Conference on Artificial Intelligence},
  volume={38},
  number={15},
  pages={16522--16530},
  year={2024}
}

@article{liDDGDataDistributionGeneration2022,
  title = {{{DDG-DA}}: {{Data Distribution Generation}} for {{Predictable Concept Drift Adaptation}}},
  shorttitle = {{{DDG-DA}}},
  author = {Li, Wendi and Yang, Xiao and Liu, Weiqing and Xia, Yingce and Bian, Jiang},
  year = {2022-06-28},
  journaltitle = {Proceedings of the AAAI Conference on Artificial Intelligence},
  volume = {36},
  number = {4},
  pages = {4092--4100},
  issn = {2374-3468},
  doi = {10.1609/aaai.v36i4.20327},
  url = {https://ojs.aaai.org/index.php/AAAI/article/view/20327},
  urldate = {2024-09-28},
  langid = {english}
}

@inproceedings{xuLearningImbalancedData2023,
  title = {Learning {{Imbalanced Data With Vision Transformers}}},
  author = {Xu, Zhengzhuo and Liu, Ruikang and Yang, Shuo and Chai, Zenghao and Yuan, Chun},
  year = {2023},
  pages = {15793--15803},
  url = {https://openaccess.thecvf.com/content/CVPR2023/html/Xu_Learning_Imbalanced_Data_With_Vision_Transformers_CVPR_2023_paper.html},
  urldate = {2024-09-29},
  booktitle = {Proceedings of the {{IEEE}}/{{CVF Conference}} on {{Computer Vision}} and {{Pattern Recognition}}},
  langid = {english}
}

@inproceedings{heMaskedAutoencodersAre2022,
  title = {Masked {{Autoencoders Are Scalable Vision Learners}}},
  author = {He, Kaiming and Chen, Xinlei and Xie, Saining and Li, Yanghao and Dollár, Piotr and Girshick, Ross},
  year = {2022},
  pages = {16000--16009},
  url = {https://openaccess.thecvf.com/content/CVPR2022/html/He_Masked_Autoencoders_Are_Scalable_Vision_Learners_CVPR_2022_paper.html},
  urldate = {2023-06-08},
  booktitle = {Proceedings of the {{IEEE}}/{{CVF Conference}} on {{Computer Vision}} and {{Pattern Recognition}}},
  langid = {english}
}

@inproceedings{touvronDeiTIIIRevenge2022,
  title = {{{DeiT III}}: {{Revenge}} of~the~{{ViT}}},
  shorttitle = {{{DeiT III}}},
  booktitle = {Computer {{Vision}} – {{ECCV}} 2022},
  author = {Touvron, Hugo and Cord, Matthieu and Jégou, Hervé},
  editor = {Avidan, Shai and Brostow, Gabriel and Cissé, Moustapha and Farinella, Giovanni Maria and Hassner, Tal},
  year = {2022},
  pages = {516--533},
  publisher = {Springer Nature Switzerland},
  location = {Cham},
  doi = {10.1007/978-3-031-20053-3_30},
  isbn = {978-3-031-20053-3},
  langid = {english}
}

@inproceedings{wangLearningRobustGlobal2019,
  title = {Learning {{Robust Global Representations}} by {{Penalizing Local Predictive Power}}},
  booktitle = {Advances in {{Neural Information Processing Systems}}},
  author = {Wang, Haohan and Ge, Songwei and Lipton, Zachary and Xing, Eric P},
  year = {2019},
  volume = {32},
  publisher = {Curran Associates, Inc.},
  url = {https://proceedings.neurips.cc/paper/2019/hash/3eefceb8087e964f89c2d59e8a249915-Abstract.html},
  urldate = {2024-11-20}
}

@inproceedings{grillBootstrapYourOwn2020,
  title = {Bootstrap {{Your Own Latent}} - {{A New Approach}} to {{Self-Supervised Learning}}},
  booktitle = {Advances in {{Neural Information Processing Systems}}},
  author = {Grill, Jean-Bastien and Strub, Florian and Altché, Florent and Tallec, Corentin and Richemond, Pierre and Buchatskaya, Elena and Doersch, Carl and Avila Pires, Bernardo and Guo, Zhaohan and Gheshlaghi Azar, Mohammad and Piot, Bilal and {kavukcuoglu}, koray and Munos, Remi and Valko, Michal},
  year = {2020},
  volume = {33},
  pages = {21271--21284},
  publisher = {Curran Associates, Inc.},
  url = {https://proceedings.neurips.cc/paper/2020/hash/f3ada80d5c4ee70142b17b8192b2958e-Abstract.html},
  urldate = {2025-01-11}
}

@inproceedings{vaswaniAttentionAllYou2017,
  title = {Attention Is {{All}} You {{Need}}},
  booktitle = {Advances in {{Neural Information Processing Systems}}},
  author = {Vaswani, Ashish and Shazeer, Noam and Parmar, Niki and Uszkoreit, Jakob and Jones, Llion and Gomez, Aidan N and Kaiser, Lukasz and Polosukhin, Illia},
  year = {2017},
  volume = {30},
  publisher = {Curran Associates, Inc.},
  url = {https://proceedings.neurips.cc/paper/2017/hash/3f5ee243547dee91fbd053c1c4a845aa-Abstract.html},
  urldate = {2023-01-06},
  langid = {american}
}

@inproceedings{chenExploringSimpleSiamese2021,
  title = {Exploring {{Simple Siamese Representation Learning}}},
  author = {Chen, Xinlei and He, Kaiming},
  year = {2021},
  pages = {15750--15758},
  url = {https://openaccess.thecvf.com/content/CVPR2021/html/Chen_Exploring_Simple_Siamese_Representation_Learning_CVPR_2021_paper.html},
  urldate = {2025-01-12},
  booktitle = {Proceedings of the {{IEEE}}/{{CVF Conference}} on {{Computer Vision}} and {{Pattern Recognition}}},
  langid = {english}
}

@article{muennighoffScalingDataConstrainedLanguage2023,
  title = {Scaling {{Data-Constrained Language Models}}},
  author = {Muennighoff, Niklas and Rush, Alexander and Barak, Boaz and Le Scao, Teven and Tazi, Nouamane and Piktus, Aleksandra and Pyysalo, Sampo and Wolf, Thomas and Raffel, Colin A.},
  year = {2023},
  journaltitle = {Advances in Neural Information Processing Systems},
  volume = {36},
  pages = {50358--50376},
  url = {https://proceedings.neurips.cc/paper_files/paper/2023/hash/9d89448b63ce1e2e8dc7af72c984c196-\\Abstract-Conference.html},
  urldate = {2025-01-13},
  langid = {english}
}

@article{hendrycks2016baseline,
  title={A baseline for detecting misclassified and out-of-distribution examples in neural networks},
  author={Hendrycks, Dan and Gimpel, Kevin},
  journal={arXiv preprint arXiv:1610.02136},
  year={2016}
}

@article{liu2020energy,
  title={Energy-based out-of-distribution detection},
  author={Liu, Weitang and Wang, Xiaoyun and Owens, John and Li, Yixuan},
  journal={Advances in neural information processing systems},
  volume={33},
  pages={21464--21475},
  year={2020}
}

@article{hendrycks2019scaling,
  title={Scaling out-of-distribution detection for real-world settings},
  author={Hendrycks, Dan and Basart, Steven and Mazeika, Mantas and Zou, Andy and Kwon, Joe and Mostajabi, Mohammadreza and Steinhardt, Jacob and Song, Dawn},
  journal={arXiv preprint arXiv:1911.11132},
  year={2019}
}

@inproceedings{wang2022vim,
  title={Vim: Out-of-distribution with virtual-logit matching},
  author={Wang, Haoqi and Li, Zhizhong and Feng, Litong and Zhang, Wayne},
  booktitle={Proceedings of the IEEE/CVF conference on computer vision and pattern recognition},
  pages={4921--4930},
  year={2022}
}

@article{sun2021react,
  title={React: Out-of-distribution detection with rectified activations},
  author={Sun, Yiyou and Guo, Chuan and Li, Yixuan},
  journal={Advances in Neural Information Processing Systems},
  volume={34},
  pages={144--157},
  year={2021}
}

@article{lee2018simple,
  title={A simple unified framework for detecting out-of-distribution samples and adversarial attacks},
  author={Lee, Kimin and Lee, Kibok and Lee, Honglak and Shin, Jinwoo},
  journal={Advances in neural information processing systems},
  volume={31},
  year={2018}
}

@inproceedings{li2023rethinking,
  title={Rethinking out-of-distribution (ood) detection: Masked image modeling is all you need},
  author={Li, Jingyao and Chen, Pengguang and He, Zexin and Yu, Shaozuo and Liu, Shu and Jia, Jiaya},
  booktitle={Proceedings of the IEEE/CVF conference on computer vision and pattern recognition},
  pages={11578--11589},
  year={2023}
}

@article{liMOODv2MaskedImage2024,
  title = {{{MOODv2}}: {{Masked Image Modeling}} for {{Out-of-Distribution Detection}}},
  shorttitle = {{{MOODv2}}},
  author = {Li, Jingyao and Chen, Pengguang and Yu, Shaozuo and Liu, Shu and Jia, Jiaya},
  year = {2024},
  journaltitle = {IEEE Transactions on Pattern Analysis and Machine Intelligence},
  volume = {46},
  number = {12},
  pages = {8994--9003},
  issn = {0162-8828},
  doi = {10.1109/TPAMI.2024.3412004},
  url = {https://www.computer.org/csdl/journal/tp/2024/12/10553645/1XH2Iao33Z6},
  urldate = {2025-01-13},
  langid = {american}
}

@inproceedings{cimpoi2014describing,
  title={Describing textures in the wild},
  author={Cimpoi, Mircea and Maji, Subhransu and Kokkinos, Iasonas and Mohamed, Sammy and Vedaldi, Andrea},
  booktitle={Proceedings of the IEEE conference on computer vision and pattern recognition},
  pages={3606--3613},
  year={2014}
}

@inproceedings{van2018inaturalist,
  title={The inaturalist species classification and detection dataset},
  author={Van Horn, Grant and Mac Aodha, Oisin and Song, Yang and Cui, Yin and Sun, Chen and Shepard, Alex and Adam, Hartwig and Perona, Pietro and Belongie, Serge},
  booktitle={Proceedings of the IEEE conference on computer vision and pattern recognition},
  pages={8769--8778},
  year={2018}
}

@inproceedings{huangMOSScalingOutofDistribution2021,
  title = {{{MOS}}: {{Towards Scaling Out-of-Distribution Detection}} for {{Large Semantic Space}}},
  shorttitle = {{{MOS}}},
  author = {Huang, Rui and Li, Yixuan},
  year = {2021},
  pages = {8710--8719},
  url = {https://openaccess.thecvf.com/content/CVPR2021/html/Huang_MOS_Towards_Scaling_Out-of-Distribution_Detection_for_Large_Semantic_Space_CVPR_2021_paper.html},
  urldate = {2025-01-16},
  booktitle = {Proceedings of the {{IEEE}}/{{CVF Conference}} on {{Computer Vision}} and {{Pattern Recognition}}},
  langid = {english}
}

@inproceedings{vanhornINaturalistSpeciesClassification2018a,
  title = {The {{INaturalist Species Classification}} and {{Detection Dataset}}},
  author = {Van Horn, Grant and Mac Aodha, Oisin and Song, Yang and Cui, Yin and Sun, Chen and Shepard, Alex and Adam, Hartwig and Perona, Pietro and Belongie, Serge},
  year = {2018},
  pages = {8769--8778},
  url = {https://openaccess.thecvf.com/content_cvpr_2018/html/Van_Horn_The_INaturalist_Species_CVPR_2018_paper.html},
  urldate = {2025-01-20},
  booktitle = {Proceedings of the {{IEEE Conference}} on {{Computer Vision}} and {{Pattern Recognition}}}
}

@inproceedings{hendrycksManyFacesRobustness2021,
  title = {The {{Many Faces}} of {{Robustness}}: {{A Critical Analysis}} of {{Out-of-Distribution Generalization}}},
  shorttitle = {The {{Many Faces}} of {{Robustness}}},
  author = {Hendrycks, Dan and Basart, Steven and Mu, Norman and Kadavath, Saurav and Wang, Frank and Dorundo, Evan and Desai, Rahul and Zhu, Tyler and Parajuli, Samyak and Guo, Mike and Song, Dawn and Steinhardt, Jacob and Gilmer, Justin},
  year = {2021},
  pages = {8340--8349},
  url = {https://openaccess.thecvf.com/content/ICCV2021/html/Hendrycks_The_Many_Faces_of_Robustness_A_Critical_Analysis_of_Out-of-Distribution_ICCV_2021_paper.html},
  urldate = {2024-11-21},
  booktitle = {Proceedings of the {{IEEE}}/{{CVF International Conference}} on {{Computer Vision}}},
  langid = {english}
}

@inproceedings{rechtImageNetClassifiersGeneralize2019,
  title = {Do {{ImageNet Classifiers Generalize}} to {{ImageNet}}?},
  booktitle = {Proceedings of the 36th {{International Conference}} on {{Machine Learning}}},
  author = {Recht, Benjamin and Roelofs, Rebecca and Schmidt, Ludwig and Shankar, Vaishaal},
  year = {2019},
  pages = {5389--5400},
  publisher = {PMLR},
  url = {https://proceedings.mlr.press/v97/recht19a.html},
  urldate = {2025-01-21},
  eventtitle = {International {{Conference}} on {{Machine Learning}}},
  langid = {english}
}

@inproceedings{hendrycksNaturalAdversarialExamples2021,
  title = {Natural {{Adversarial Examples}}},
  author = {Hendrycks, Dan and Zhao, Kevin and Basart, Steven and Steinhardt, Jacob and Song, Dawn},
  year = {2021},
  pages = {15262--15271},
  url = {https://openaccess.thecvf.com/content/CVPR2021/html/Hendrycks_Natural_Adversarial_Examples_CVPR_2021_paper.html},
  urldate = {2024-11-21},
  eventtitle = {Proceedings of the {{IEEE}}/{{CVF Conference}} on {{Computer Vision}} and {{Pattern Recognition}}},
  langid = {english}
}

@inproceedings{wangViMOutofDistributionVirtualLogit2022,
  title = {{{ViM}}: {{Out-of-Distribution With Virtual-Logit Matching}}},
  shorttitle = {{{ViM}}},
  author = {Wang, Haoqi and Li, Zhizhong and Feng, Litong and Zhang, Wayne},
  year = {2022},
  pages = {4921--4930},
  url = {https://openaccess.thecvf.com/content/CVPR2022/html/Wang_ViM_Out-of-Distribution_With_Virtual-Logit_Matching_CVPR_2022_paper.html},
  urldate = {2025-01-21},
  eventtitle = {Proceedings of the {{IEEE}}/{{CVF Conference}} on {{Computer Vision}} and {{Pattern Recognition}}},
  langid = {english}
}

@online{ridnikImageNet21KPretrainingMasses2021,
  title = {{{ImageNet-21K Pretraining}} for the {{Masses}}},
  author = {Ridnik, Tal and Ben-Baruch, Emanuel and Noy, Asaf and Zelnik-Manor, Lihi},
  year = {2021},
  eprint = {2104.10972},
  eprinttype = {arXiv},
  doi = {10.48550/arXiv.2104.10972},
  url = {http://arxiv.org/abs/2104.10972},
  urldate = {2025-01-21},
  pubstate = {prepublished}
}

@article{yuDetectingGroupConcept2023,
  title = {Detecting Group Concept Drift from Multiple Data Streams},
  author = {Yu, Hang and Liu, Weixu and Lu, Jie and Wen, Yimin and Luo, Xiangfeng and Zhang, Guangquan},
  year = {2023},
  journaltitle = {Pattern Recognition},
  shortjournal = {Pattern Recognition},
  volume = {134},
  pages = {109113},
  issn = {0031-3203},
  doi = {10.1016/j.patcog.2022.109113},
  url = {https://www.sciencedirect.com/science/article/pii/S0031320322005933},
  urldate = {2024-04-26}
}

@article{wangSelfadaptiveEnsembleUser2024,
  title = {A Self-Adaptive Ensemble for User Interest Drift Learning},
  author = {Wang, Kun and Xiong, Li and Liu, Anjin and Zhang, Guangquan and Lu, Jie},
  year = {2024},
  journaltitle = {Neurocomputing},
  shortjournal = {Neurocomputing},
  volume = {577},
  pages = {127308},
  issn = {0925-2312},
  doi = {10.1016/j.neucom.2024.127308},
  url = {https://www.sciencedirect.com/science/article/pii/S0925231224000791},
  urldate = {2024-02-21}
}

@article{jiaoDynamicEnsembleSelection2024,
  title = {Dynamic {{Ensemble Selection}} for {{Imbalanced Data Streams With Concept Drift}}},
  author = {Jiao, Botao and Guo, Yinan and Gong, Dunwei and Chen, Qiuju},
  year = {2024},
  journaltitle = {IEEE Transactions on Neural Networks and Learning Systems},
  volume = {35},
  number = {1},
  pages = {1278--1291},
  issn = {2162-2388},
  doi = {10.1109/TNNLS.2022.3183120},
  url = {https://ieeexplore.ieee.org/abstract/document/9802893},
  urldate = {2024-05-20}
}

@article{cerqueiraSTUDDStudentTeacher2023,
  title = {{{STUDD}}: A Student–Teacher Method for Unsupervised Concept Drift Detection},
  shorttitle = {{{STUDD}}},
  author = {Cerqueira, Vitor and Gomes, Heitor Murilo and Bifet, Albert and Torgo, Luis},
  year = {2023},
  journaltitle = {Machine Learning},
  shortjournal = {Mach Learn},
  volume = {112},
  number = {11},
  pages = {4351--4378},
  issn = {1573-0565},
  doi = {10.1007/s10994-022-06188-7},
  url = {https://doi.org/10.1007/s10994-022-06188-7},
  urldate = {2024-03-04},
  langid = {english}
}

@article{yuLearnadaptConceptDrift2022,
  title = {Learn-to-Adapt: {{Concept}} Drift Adaptation for Hybrid Multiple Streams},
  shorttitle = {Learn-to-Adapt},
  author = {Yu, En and Song, Yiliao and Zhang, Guangquan and Lu, Jie},
  year = {2022},
  journaltitle = {Neurocomputing},
  shortjournal = {Neurocomputing},
  volume = {496},
  pages = {121--130},
  issn = {0925-2312},
  doi = {10.1016/j.neucom.2022.05.025},
  url = {https://www.sciencedirect.com/science/article/pii/S0925231222005550},
  urldate = {2024-03-04}
}

@inproceedings{dengComprehensiveKnowledgeDistillation2021,
  title = {Comprehensive {{Knowledge Distillation}} with {{Causal Intervention}}},
  booktitle = {Advances in {{Neural Information Processing Systems}}},
  author = {Deng, Xiang and Zhang, Zhongfei},
  year = {2021},
  volume = {34},
  pages = {22158--22170},
  publisher = {Curran Associates, Inc.},
  url = {https://proceedings.neurips.cc/paper/2021/hash/b9f35816f460ab999cbc168c4da26ff3-Abstract.html},
  urldate = {2024-12-13},
  langid = {american}
}

@online{zhangCausalWalkDebiasing2024,
  title = {Causal {{Walk}}: {{Debiasing Multi-Hop Fact Verification}} with {{Front-Door Adjustment}}},
  shorttitle = {Causal {{Walk}}},
  author = {Zhang, Congzhi and Zhang, Linhai and Zhou, Deyu},
  year = {2024},
  eprint = {2403.02698},
  eprinttype = {arXiv},
  doi = {10.48550/arXiv.2403.02698},
  url = {http://arxiv.org/abs/2403.02698},
  urldate = {2025-01-09},
  langid = {american},
  pubstate = {prepublished}
}

@inproceedings{tangLongTailedClassificationKeeping2021,
  title = {Long-{{Tailed Classification}} by {{Keeping}} the {{Good}} and {{Removing}} the {{Bad Momentum Causal Effect}}},
  booktitle = {Advances in {{Neural Information Processing Systems}}},
  author = {Tang, Kaihua and Huang, Jianqiang and Zhang, Hanwang},
  year = {2021},
  url = {https://proceedings.neurips.cc//paper/2020/file/1091660f3dff84fd648efe31391c5524-Paper.pdf},
  urldate = {2021-06-09},
  eventtitle = {Advances in {{Neural Information Processing Systems}}},
  langid = {english}
}

@inproceedings{wangDenseContrastiveLearning2021a,
  title = {Dense {{Contrastive Learning}} for {{Self-Supervised Visual Pre-Training}}},
  author = {Wang, Xinlong and Zhang, Rufeng and Shen, Chunhua and Kong, Tao and Li, Lei},
  year = {2021},
  pages = {3024--3033},
  url = {https://openaccess.thecvf.com/content/CVPR2021/html/Wang_Dense_Contrastive_Learning_for_Self-Supervised_Visual_Pre-Training_CVPR_2021_paper.html},
  urldate = {2025-01-29},
  booktitle = {Proceedings of the {{IEEE}}/{{CVF Conference}} on {{Computer Vision}} and {{Pattern Recognition}}},
  langid = {english}
}

@article{lu2020data,
  title={Data-driven decision support under concept drift in streamed big data},
  author={Lu, Jie and Liu, Anjin and Song, Yiliao and Zhang, Guangquan},
  journal={Complex \& intelligent systems},
  volume={6},
  number={1},
  pages={157--163},
  year={2020},
  publisher={Springer}
}

@inproceedings{chenimportance,
  title={On the Importance and Applicability of Pre-Training for Federated Learning},
  author={Chen, Hong-You and Tu, Cheng-Hao and Li, Ziwei and Shen, Han Wei and Chao, Wei-Lun},
  year = {2023},
  booktitle={The Eleventh International Conference on Learning Representations}
}

@inproceedings{panchal2023flash,
  title={Flash: Concept drift adaptation in federated learning},
  author={Panchal, Kunjal and Choudhary, Sunav and Mitra, Subrata and Mukherjee, Koyel and Sarkhel, Somdeb and Mitra, Saayan and Guan, Hui},
  booktitle={International Conference on Machine Learning},
  pages={26931--26962},
  year={2023},
  organization={PMLR}
}

@inproceedings{
garg2024ticclip,
title={TiC-{CLIP}: Continual Training of {CLIP} Models},
author={Saurabh Garg and Mehrdad Farajtabar and Hadi Pouransari and Raviteja Vemulapalli and Sachin Mehta and Oncel Tuzel and Vaishaal Shankar and Fartash Faghri},
booktitle={The Twelfth International Conference on Learning Representations},
year={2024},
url={https://openreview.net/forum?id=TLADT8Wrhn}
}

@inproceedings{gao2022feddc,
  title={Feddc: Federated learning with non-iid data via local drift decoupling and correction},
  author={Gao, Liang and Fu, Huazhu and Li, Li and Chen, Yingwen and Xu, Ming and Xu, Cheng-Zhong},
  booktitle={Proceedings of the IEEE/CVF conference on computer vision and pattern recognition},
  pages={10112--10121},
  year={2022}
}

@inproceedings{jiang2022harmofl,
  title={Harmofl: Harmonizing local and global drifts in federated learning on heterogeneous medical images},
  author={Jiang, Meirui and Wang, Zirui and Dou, Qi},
  booktitle={Proceedings of the AAAI Conference on Artificial Intelligence},
  volume={36},
  number={1},
  pages={1087--1095},
  year={2022}
}

@inproceedings{zheng2023autofed,
  title={Autofed: Heterogeneity-aware federated multimodal learning for robust autonomous driving},
  author={Zheng, Tianyue and Li, Ang and Chen, Zhe and Wang, Hongbo and Luo, Jun},
  booktitle={Proceedings of the 29th annual international conference on mobile computing and networking},
  pages={1--15},
  year={2023}
}

@article{liang2022effective,
  title={Effective adaptation in multi-task co-training for unified autonomous driving},
  author={Liang, Xiwen and Wu, Yangxin and Han, Jianhua and Xu, Hang and Xu, Chunjing and Liang, Xiaodan},
  journal={Advances in Neural Information Processing Systems},
  volume={35},
  pages={19645--19658},
  year={2022}
}

@inproceedings{mitrovicrepresentation,
  title={Representation Learning via Invariant Causal Mechanisms},
  author={Mitrovic, Jovana and McWilliams, Brian and Walker, Jacob C and Buesing, Lars Holger and Blundell, Charles},
  booktitle={International Conference on Learning Representations},
    year = {2021}
}

@article{yu2025drift,
  title={Drift-aware collaborative assistance mixture of experts for heterogeneous multistream learning},
  author={Yu, En and Lu, Jie and Wang, Kun and Yang, Xiaoyu and Zhang, Guangquan},
  journal={arXiv preprint arXiv:2508.01598},
  year={2025}
}

@article{yang2025walking,
  title={Walking the tightrope: Disentangling beneficial and detrimental drifts in non-stationary custom-tuning},
  author={Yang, Xiaoyu and Lu, Jie and Yu, En},
  journal={arXiv preprint arXiv:2505.13081},
  year={2025}
}

@article{yang2024adaptingmultimodallargelanguage,
  title={Adapting Multi-modal Large Language Model to Concept Drift From Pre-training Onwards}, 
  author={Yang, Xiaoyu and Lu, Jie and Yu, En},
  journal={arXiv preprint arXiv:2405.13459},
  year={2024}
}

@article{chen2024general,
  title={A general variation-driven network for medical image synthesis},
  author={Chen, Yufei and Yang, Xiaoyu and Yue, Xiaodong and Lin, Xiang and Zhang, Qi and Fujita, Hamido},
  journal={Applied Intelligence},
  volume={54},
  number={4},
  pages={3295--3307},
  year={2024},
  publisher={Springer Nature BV}
}

@article{yang2022local,
  title={Local linear embedding based interpolation neural network in pancreatic tumor segmentation},
  author={Yang, Xiaoyu and Chen, Yufei and Yue, Xiaodong and Ma, Chao and Yang, Panpan},
  journal={Applied Intelligence},
  volume={52},
  number={8},
  pages={8746--8756},
  year={2022},
  publisher={Springer}
}

@article{yang2024segmentation,
  title={Segmentation and vascular vectorization for coronary artery by geometry-based cascaded neural network},
  author={Yang, Xiaoyu and Xu, Lijian and Yu, Simon and Xia, Qing and Li, Hongsheng and Zhang, Shaoting},
  journal={IEEE Transactions on Medical Imaging},
  year={2024},
  publisher={IEEE}
}

@article{yangTdistributedSphericalFeature2023,
  title = {T-Distributed {{Spherical Feature Representation}} for {{Imbalanced Classification}}},
  author = {Yang, Xiaoyu and Chen, Yufei and Yue, Xiaodong and Xu, Shaoxun and Ma, Chao},
  year = {2023},
  journal = {Proceedings of the AAAI Conference on Artificial Intelligence},
  volume = {37},
  year = {2023},
  number = {9},
  pages = {10825--10833},
  issn = {2374-3468},
  doi = {10.1609/aaai.v37i9.26284},
  url = {https://ojs.aaai.org/index.php/AAAI/article/view/26284},
  urldate = {2024-03-06},
  issue = {9},
  langid = {english}
}

@article{yang2024enhancing,
  title={Enhancing Visual Grounding and Generalization: A Multi-Task Cycle Training Approach for Vision-Language Models}, 
  author={Xiaoyu Yang and Lijian Xu and Hao Sun and Hongsheng Li and Shaoting Zhang},
  year={2024},
  journal={arXiv preprint arXiv:2311.12327},
}

@article{yangMaskedImageContrastive2024,
  title={One Leaf Reveals the Season: Occlusion-Based Contrastive Learning with Semantic-Aware Views for Efficient Visual Representation},
  author={Yang, Xiaoyu and Xu, Lijian and Li, Hongsheng and Zhang, Shaoting},
  journal={arXiv preprint arXiv:2411.09858},
  year={2024}
}

@article{yang2025learning,
  title={Learning from All: Concept Alignment for Autonomous Distillation from Multiple Drifting MLLMs},
  author={Yang, Xiaoyu and Lu, Jie and Yu, En},
  journal={arXiv preprint arXiv:2510.04142},
  year={2025}
}

@inproceedings{yang2021variational,
  title={Variational synthesis network for generating micro computed tomography from cone beam computed tomography},
  author={Yang, Xiaoyu and Chen, Yufei and Yue, Xiaodong and Lin, Xiang and Zhang, Qi},
  booktitle={2021 IEEE International Conference on Bioinformatics and Biomedicine (BIBM)},
  pages={1611--1614},
  year={2021},
  organization={IEEE}
}
}


\clearpage
\setcounter{page}{1}
\maketitlesupplementary

\section{Related Works}
\label{appendix:relatedwork}

\subsection{Concept Drift}

In the comprehensive survey conducted by Lu et al. \cite{luLearningConceptDrift2019, lu2020data}, existing approaches for concept drift handling are systematically classified into three primary categories: error rate-based methods \cite{wangSelfadaptiveEnsembleUser2024, jiaoDynamicEnsembleSelection2024}, data distribution-based methods \cite{yang2024adaptingmultimodallargelanguage,cerqueiraSTUDDStudentTeacher2023}, and multiple hypothesis-based techniques \cite{yu2024online, yuLearnadaptConceptDrift2022, yu2025drift}. Our proposed methodology falls under the category of distribution-based concept drift adaptation approaches. These distribution-driven techniques distinguish themselves by not only enabling precise drift identification through explicit statistical distribution analysis but also providing a multidimensional characterization of drift patterns - including temporal occurrence detection, affected feature space localization, and quantitative severity assessment. This dual capability of detection coupled with comprehensive drift diagnostics establishes distribution-based methods as particularly valuable for developing adaptive systems that require interpretable drift understanding and targeted model adjustments.

Besides, Online Boosting Adaptive Learning (OBAL) method \cite{yu2024online} is proposed to address the challenges of concept drift and negative transfer in multistream classification. OBAL employs a dual-phase approach: an initial model is built using the Adaptive Covariate Shift Adaptation (AdaCOSA) algorithm to handle covariate shifts and learn dynamic correlations among streams. In the online phase, a Gaussian Mixture Model-based weighting mechanism is integrated to manage asynchronous drift. Meanwhile, CDMLLM \cite{yang2024adaptingmultimodallargelanguage} reveals that vision-language models suffer significant bias from concept drift during pre-training and fine-tuning. To address this, the authors propose a unified concept drift framework integrating T-distribution-based adaptation for long-tailed calibration and explicit OOD detection, demonstrating enhanced robustness in open-world multi-modal alignment through systematic distribution modelling. Beyond on drift detection in single data stream, GDDM \cite{yuDetectingGroupConcept2023} focuses on addressing group concept drift across multiple data streams, where individual drifts may go undetected due to subtle changes in underlying distributions. The proposed method introduces a distribution-free test statistic to detect concept drift in these complex scenarios. By designing an online learning algorithm for streaming data, the approach accurately identifies concept drift caused by hypothesis testing. Beyond that, DDG-DA \cite{liDDGDataDistributionGeneration2022} proactively models predictable factors influencing environmental evolution. The approach involves training a predictor to estimate future data distribution, using this information to generate training samples, and then training models on the generated data. By leveraging predictable factors to forecast data distribution changes, DDG-DA aims to enhance model performance in handling concept drift in streaming data. Furthermore, STUDD\cite{cerqueiraSTUDDStudentTeacher2023} proposes a teacher-student paradigm to enable unsupervised drift detection through deviation analysis of their predictive consistency. This approach leverages model disagreement as a proxy signal, bypassing dependency on ground-truth labels during deployment while maintaining detection sensitivity.

\subsection{Causal Inference}

Recently, increasing researchers have incorporated causal inference into deep-learning models, especially in large models. Deconfounded Image Captioning (DIC) \cite{yangDeconfoundedImageCaptioning2023} is proposed to address dataset bias in vision-language models through a causal lens, that integrates backdoor and front-door adjustments for systematic bias mitigation. The framework provides principled causal analysis of spurious correlations in multimodal alignment, offering theoretical grounding for decomposing bias sources through structured interventions. Likewise, aiming for spurious correlations induced by visual and linguistic biases during training, CIIC \cite{liuShowDeconfoundTell2022} is proposed as a causal intervention framework combining an Interventional Object Detector (IOD) and Interventional Transformer Decoder (ITD) guided by structural causal models. By applying backdoor adjustment through IOD's feature disentanglement and ITD's dual de-confounding mechanism, their approach systematically mitigates confounding effects across encoding and decoding stages, demonstrating enhanced generalization through causal correlation modeling. Similarly, targeting multi-hop fact verification bias in the large language model, Causal Walk \cite{zhangCausalWalkDebiasing2024} is proposed, a front-door adjustment framework that disentangles complex spurious correlations in evidence chains. 
The method models reasoning paths as mediators in structural causal models, decomposing causal effects via random walk-based treatment-mediator estimation and geometric mean-based mediator-outcome approximation. By integrating adversarial and symmetric datasets synthesized with large language models, the approach demonstrates superior debiasing performance. 

Additionally, causal inference is widely used in representation learning. Comprehensive Interventional Distillation (CID) \cite{dengComprehensiveKnowledgeDistillation2021} integrates causal intervention with class-aware representation alignment. By reinterpreting teacher logits as contextual confounders and applying counterfactual pruning through structural causal models, CID systematically disentangles beneficial semantic patterns from dataset-specific biases. This approach demonstrates enhanced generalization through bias-invariant knowledge transfer. Besides, De-confound-TDE \cite{tangLongTailedClassificationKeeping2021} establishes a causal framework for long-tailed classification, identifying SGD momentum as a paradoxical confounder that simultaneously harms tail-class predictions while benefiting representation learning. Through causal intervention during training and counterfactual reasoning at inference, the method disentangles momentum’s detrimental bias from its beneficial mediation effects. Meanwhile, CCM \cite{miaoDomainGeneralizationContrastive2022} is proposed to address domain generalization through causal invariance principles. The framework integrates front-door adjustment with contrastive learning to quantify stable causal effects across domains, explicitly modeling domain shifts via a three-stage process: domain-conditioned supervision for feature correlation, causal effect measurement through structured path manipulation, and contrastive clustering for class-consistent representations. Similarly, CIRL \cite{lvCausalityInspiredRepresentation2022} advances domain generalization through causal factorization, proposing a structural causal model that decomposes inputs into invariant causal mechanisms and domain-specific non-causal factors. It enforces three critical properties: causal/non-causal separation, joint independence, and causal sufficiency for classification. 

Besides, C2L \cite{choiC2LCausallyContrastive2022} addresses model fragility to spurious patterns through contrastive counterfactual synthesis, proposing a collective decision framework that aggregates predictions across generated counterfactual sets. Unlike conventional augmentation limited by dataset-inherent biases, this approach probabilistically supervises causal invariance through distributional consensus, demonstrating enhanced robustness against attribution bias and domain shifts. Furthermore, ABCD \cite{rohekarCausalInterpretationSelfAttention2023} establishes a causal interpretation of Transformer self-attention mechanisms, modeling them as structural equation estimators that capture conditional independence relations through partial correlation analysis in deep attention layers. This framework enables zero-shot causal discovery over input sequences while accounting for latent confounders, effectively repurposing pre-trained models for causal graph inference. What’s more, Causal Attention (CATT) \cite{yangCausalAttentionVisionLanguage2021} implements front-door adjustment to address confounding bias in attention mechanisms via dual-path processing of In-Sample and Cross-Sample Attention. By forcibly integrating external sample contexts through CS-ATT while maintaining standard attention conventions, CATT dynamically mitigates spurious correlations without requiring confounder specification. 

\subsection{Contrastive Pre-training}

The seminal work of SimCLR \cite{chenSimpleFrameworkContrastive2020} establishes foundational principles for contrastive pretraining through instance discrimination objectives and systematic augmentation strategies. While achieving remarkable performance in static environments, its reliance on fixed augmentation policies and stationary negative sampling assumes temporal consistency of latent patterns – an assumption violated under concept drift scenarios. Nevertheless, SimCLR's demonstration of invariant representation learning through normalized temperature-scaled loss provides crucial architectural groundwork for developing drift-aware contrastive frameworks, particularly in modeling evolving feature relationships through dynamic positive/negative pair formulation.

Besides, the MoCo framework \cite{heMomentumContrastUnsupervised2020} and its subsequent evolution MoCo v3 \cite{chenEmpiricalStudyTraining2021} establish critical momentum-based mechanisms for contrastive learning through dynamic memory banks and stabilized key encoder updates. It is primarily designed to maintain consistency in negative sample maintenance and gradient stability. However, their fixed momentum schedules and stationary target assumptions limit adaptability to abrupt distribution shifts as we discussed aforementioned. Likewise, BYOL \cite{grillBootstrapYourOwn2020} employs two neural networks—an online network and a target network—that interact and learn from each other. Specifically, the online network is trained to predict the representation of an augmented image view produced by the target network, while the target network is updated using a slow-moving average of the online network's parameters. Similarly, the SiamSim framework \cite{chenExploringSimpleSiamese2021} advances contrastive representation learning through dynamic similarity calibration, employing a dual-path Siamese architecture with momentum-aligned feature projectors. 

Meanwhile, DINO \cite{caronEmergingPropertiesSelfSupervised2021} and DINO v2 \cite{zhouDINOWMWorldModels2024} advance self-distillation paradigms through momentum-based teacher-student mechanisms, leveraging global-local attention consistency in Vision Transformers to learn semantically structured representations. While achieving state-of-the-art in static self-supervised learning, their fixed teacher-student update schedules and monolithic prototype banks implicitly assume stationarity of feature distributions. DINO v2's introduction of partitioned expertise through specialized sub-networks demonstrates partial adaptability to data variations, suggesting pathways for concept drift mitigation through dynamic expert routing. These works collectively establish critical baselines for stability-aware distillation architectures in non-stationary pre-training scenarios.

Furthermore, DenseCL \cite{wangDenseContrastiveLearning2021a}  advances contrastive learning through localized feature alignment, introducing dense instance discrimination that operates at both image-level and pixel-level granularity. By enforcing spatial consistency through region-to-region contrastive pretext tasks, the method enhances model sensitivity to fine-grained visual patterns. While primarily designed for dense prediction tasks, its hierarchical contrast mechanism demonstrates the importance of multi-scale feature stabilization.

While contemporary contrastive pretraining methods achieve remarkable performance under static data distributions, their reliance on closed-world stationarity assumptions presents a fundamental limitation: they inherently lack mechanisms for concept drift adaptation when deployed in non-stationary environments with evolving data streams.

\section{Simple and Efficient Implementation with Scalability Advantages}
\label{appendxi:imple}

Our RCP method is designed for simplicity and efficiency, integrating seamlessly with existing contrastive learning pipelines. RCP requires only minimal modifications to standard architectures like MoCo v3~\cite{chenEmpiricalStudyTraining2021}. The primary adjustments involve incorporating our intervention module, which operates on the feature embeddings obtained from the student and momentum encoders. To support the larger effective batch sizes beneficial for capturing broader drift adaptation windows, we also leverage a masked contrastive strategy, inspired by recent work~\cite{yangMaskedImageContrastive2024,yang2024enhancing}.

Furthermore, RCP offers exceptional scalability advantages with respect to both data and model size. Firstly, it robustly accommodates diverse and expanding datasets during pre-training. By mitigating the adverse effects of concept drift, RCP ensures that model performance consistently benefits from increased data volume and variety, aligning with established scaling laws~\cite{muennighoffScalingDataConstrainedLanguage2023} without succumbing to drift-induced degradation. Secondly, RCP enhances the training stability of larger models in dynamic data environments. This improved stability facilitates the pre-training of models with significantly more parameters and deeper architectures, pushing the boundaries of what can be learned from evolving data streams.




\section{Experiments}

In this section, we first introduce the details of utilized datasets for pre-training and various downstream tasks. Subsequently, implementation details are provided. 

\subsection{Datasets}

ImageNet-LT \cite{liuLargeScaleLongTailedRecognition2019} is a long-tailed subset derived from the ImageNet \cite{russakovskyImagenetLargeScale2015} dataset, designed to benchmark machine learning models under realistic class imbalance. It contains 115k training images across 1,000 categories, with class frequencies ranging from 1,280 (head classes) to merely 5 (tailed classes), simulating real-world data skewness. The validation set is balanced (20 images per class), while the test set aligns with the standard ImageNet validation data (50,000 images). It serves as a pivotal resource for studying long-tail recognition challenges, including few-shot learning, open-set generalization, and algorithmic fairness, while retaining compatibility with the original ImageNet hierarchy and evaluation protocols.

iNaturalist2018 \cite{vanhornINaturalistSpeciesClassification2018a} is a large-scale dataset designed to study fine-grained recognition under naturally occurring long-tailed class distributions. It comprises 675k training and validation images spanning 5,089 species hierarchically organized into 13 super-categories, with severe imbalance reflecting ecological rarity. The dataset integrates taxonomic metadata, geospatial coordinates, and temporal annotations, supporting research in hierarchical learning, transfer learning, and self-supervised pretraining. 


In terms of OOD detection, the Texture (DTD) \cite{cimpoi2014describing} is a comprehensive collection of 5,640 texture images, meticulously organized into 47 categories based on human-centric perceptual attributes. Designed to support texture analysis and recognition tasks, DTD is divided into training, validation, and test sets, each containing 40 images per category. Besides, the iNat-OOD dataset \cite{huangMOSScalingOutofDistribution2021} is a specialized subset of the iNaturalist  \cite{vanhornINaturalistSpeciesClassification2018a} dataset, designed for evaluating out-of-distribution (OOD) detection methods in large-scale image classification tasks. The dataset is particularly valuable for testing the robustness of models in identifying novel or unseen categories, especially in ecological and biodiversity contexts. Meanwhile, ImageNet-O \cite{hendrycksNaturalAdversarialExamples2021}  is a specialized dataset designed to evaluate the robustness of visual models in detecting out-of-distribution (OOD) samples. It consists of 2,000 images from classes not included in the standard ImageNet-1k \cite{russakovskyImagenetLargeScale2015}  dataset, making it a valuable benchmark for testing OOD detection methods. Additionally, OpenImage-O \cite{wangViMOutofDistributionVirtualLogit2022}is a large-scale, manually annotated dataset designed for out-of-distribution (OOD) detection tasks. It is derived from the OpenImage-V3 test set, which contains a diverse and natural distribution of images collected from Flickr without predefined class names or tags. This dataset aims to overcome the limitations of existing OOD benchmarks by providing a more realistic and challenging testbed for evaluating the robustness of computer vision models


Regarding domain shift, ImageNet-V2 \cite{rechtImageNetClassifiersGeneralize2019} is a comprehensive test set designed to evaluate the robustness and generalization of image classification models. It comprises 10,000 images, with 10 images per class, closely following the original labeling protocol of ImageNet. This dataset is instrumental in assessing how well models trained on the original ImageNet dataset can generalize to new, unseen data, making it a valuable resource for advancing computer vision research. The ImageNet-Sketch \cite{wangLearningRobustGlobal2019} is a unique collection of 50,000 hand-drawn sketch images, with 50 images for each of the 1,000 ImageNet classes. Constructed using Google Image queries, this dataset is designed to evaluate models' ability to learn out-of-domain semantics at the ImageNet scale. ImageNet-R \cite{hendrycksManyFacesRobustness2021} is a collection of 30k images representing 200 ImageNet \cite{russakovskyImagenetLargeScale2015}  classes. These images are artistic renditions, including art, cartoons, graffiti, embroidery, origami, paintings, and other forms of creative expressions. The dataset is designed to test the robustness and generalization capabilities of image classification models when faced with non-standard inputs.
\begin{table}[tbp]
    \caption{The pre-training hyperparameters.}
    \label{table:pretrain}
    \centering
        \begin{tabular}{@{}lccc@{}}
        \toprule
                            & ViT-S/16    & ViT-B/16    & ViT-L/16    \\ \midrule
        Training Epochs     & 800         & 800         & 800         \\
        Warmup Epochs       & 40          & 40          & 40          \\
        Optimizer           & AdamW       & AdamW       & AdamW       \\
        Base Learning Rate  & 1.5e-4      & 1.5e-4      & 1.5e-4      \\
        Learning Rate Decay & Cosine      & Cosine      & Cosine      \\
        Adam $\beta$        & (0.9, 0.95) & (0.9, 0.95) & (0.9, 0.95) \\
        Weight Decay        & 0.05        & 0.05        & 0.05        \\
        Eff. Batch Size     & 32,000      & 9,600       & 2,400       \\ \bottomrule
        \end{tabular}
\end{table}

\begin{table}[htbp]
\caption{The hyperparameters of fine-tuning and linear probing.}
\label{table:hyper-ds}
\centering
\begin{tabular}{@{}lccc@{}}
\toprule
Downstream Tasks    & \multicolumn{3}{c}{Fine-tuning}    \\
Models              & ViT-S/16  & ViT-B/16 & ViT-L/16    \\ \midrule
Training Epochs     & 100       & 100      & 50          \\
Warmup Epochs       & 5         & 5        & 5           \\
Optimizer           & AdamW     & AdamW    & AdamW       \\
Base Learning Rate  & 5e-4      & 5e-4     & 1e-3        \\
Learning Rate Decay & Cosine    & Cosine   & Cosine      \\
Weight Decay        & 0.05      & 0.05     & 0.05        \\
Eff. Batch Size     & 1,024     & 1,024    & 1,024       \\ \midrule

Downstream Tasks    & \multicolumn{3}{c}{Linear Probing} \\
Models              & ViT-S/16   & ViT-B/16  & ViT-L/16  \\ \midrule
Training Epochs     & 90         & 90        & 50        \\
Warmup Epochs       & 10         & 10        & 10        \\
Optimizer           & LARS       & LARS      & LARS      \\
Base Learning Rate  & 0.1        & 0.1       & 0.1       \\
Learning Rate Decay & Cosine     & Cosine    & Cosine    \\
Weight Decay        & 0.0        & 0.0       & 0.0       \\
Eff. Batch Size     & 8,192      & 8,192     & 1,024     \\ \bottomrule

\end{tabular}
\end{table}
\subsection{Implementation Details}
\label{appendix:expdetails}
We employ ViT-Small, ViT-Base and ViT-Large as our visual backbones, respectively. Among them, Vit-Base consists of 12 transformer encoder layers and an FFN intermediate size of 3,072. The hidden dimensions of the ViT-Base are 768, with 12 attention heads. The number of parameters is about 86 million. The input image size is set to $224\times 224$. For ViT-Small, ViT-S/16 comprises 12 transformer encoder layers with an FFN intermediate size of 1,536. The input image resolution is maintained at $224\times 224$, utilizing a patch size of $16\times 16$. The hidden dimension of ViT-Small is 384, featuring 6 parallel attention heads. The total parameter count approximates 22 million. In terms of the ViT-Large, ViT-L/16 consists of 24 transformer encoder layers and an FFN intermediate size of 4,096. The input image size is set to $224\times 224$, with a patch size of $16\times 16$. The hidden dimensions of the ViT-Large are 1,024, with 16 attention heads. And, the number of parameters is about 307 million.

In terms of the pre-training progress, the hyperparameters are presented in Table \ref{table:pretrain}. 
We utilize the AdamW optimizer, which is configured with a cosine annealing schedule as the learning policy. The initial learning rate is set to $2\times10^{-5}$, and the AdamW optimizer is employed with hyperparameters $\beta= (0.9, 0.98)$. Additionally, we set the weight decay to 0.05 and the dropout rate to 0.1. During the first 40 warm-up epochs, the learning rate increases to $1.5\times10^{-4}$, and subsequently decays to $10^{-7}$. Unless otherwise specified, the pre-training of our resilient contrastive model consists of 800, executed on $2\times 2$ NVIDIA A100 GPUs.

While in the fine-tuning and linear probing on downstream task of classification, the hyperparameters are exhibited in Table \ref{table:hyper-ds}. In the fine-tuning, the initial learning rate is $5\times10^{-4}$ in ViT-S/16 and ViT-B/16, while $10^{-3}$ in ViT-L/16. Likewise, ViT-S/16 and ViT-B/16 need more training iterations with 100 epochs while 50 epochs within ViT-L/16, which are executed on $2\times 2$ NVIDIA A100 GPUs.

\subsection{Additional Experiments}

Additionally, we also conducted zero-shot experiments on TIC-DataComp(L) to verify our methods, following the settings of TiC-CLIP\cite{garg2024ticclip}. Our methods achieve superior results compared to other methods on TiC-CLIP datasets. Furthermore, it is noted that compared methods are based on vision-language models with image-text pairs, while we only used images for testing ViT.

\begin{table}[htbp]
\centering
\begin{tabular}{@{}lccc@{}}
\toprule
                 & ImageNet                             & \begin{tabular}[c]{@{}c@{}}ImageNet\\      dist shift\end{tabular} & Flickr30k                            \\ \midrule
Sequential       & 44.7                                 & 37.4                                                               & 48.4                                 \\
Patching         & 45.8                                 & 38.9                                                               & 49.7                                 \\
Cumulative-Exp   & 47.3                                 & 39.6                                                               & 50.8                                 \\
Cumulative-Equal & 47.7                                 & 40.3                                                               & 51.8                                 \\
Cumulative-All   & 53.0                                   & 44.3                                                               & 54.4                                 \\
Oracle           & 53.6                                 & 44                                                                 & 53.9                                 \\
\rowcolor[HTML]{EBEBEB} 
\textbf{Ours}    & {\color[HTML]{FE0000} \textbf{54.1}} & {\color[HTML]{FE0000} \textbf{45.9}}                               & {\color[HTML]{FE0000} \textbf{54.6}} \\ \bottomrule
\end{tabular}
\caption{Zero-shot experiments on TIC-DataComp(L).}
\label{tab:my-table}
\end{table}

Moreover, we have added more experimental results of DINO with our proposed module to demonstrate the generalization of our module on various contrastive frameworks. The results show that our proposed module can also improve the performance of DINO under non-stationary drift. It demonstrates that our proposed intervention itself is agnostic to the backbone or the specific contrastive framework: any model that maintains a momentum-updated teacher and processes mini-batches whose distribution may drift can incorporate our proposed module.

\begin{table}[htbp]
\centering
\begin{tabular}{@{}lllll@{}}
\toprule
             & Many & Medium & Few  & All  \\ \midrule
\multicolumn{5}{l}{ImageNet-LT}            \\
DINO         & 64.8 & 35.5   & 11.3 & 43.3 \\
DINO+Ours    & 67.1 & 37.7   & 14.9 & 45.9 \\
MoCo v3      & 70.8 & 40.7   & 14.3 & 48.5 \\
MoCo v3+Ours & 72.2 & 43.1   & 16   & 50.3 \\ \midrule
\multicolumn{5}{l}{iNaturalist2018}        \\
DINO         & 76.7 & 67.1   & 61.2 & 65.5 \\
DINO+Ours    & 77.1 & 67.5   & 62.1 & 66.1 \\
MoCo v3      & 76.6 & 65.8   & 62.8 & 65.6 \\
MoCo v3+Ours & 77.7 & 68.7   & 63.8 & 67.5 \\ \bottomrule
\end{tabular}
\caption{Generalization of our module on various contrastive frameworks.}
\label{tab:my-table}
\end{table}

\end{document}